\documentclass{article}
\usepackage[nonatbib,final]{sty/neurips_2024}       
\usepackage{sty/mymath}
\usepackage[utf8]{inputenc} 
\usepackage[T1]{fontenc}    
\usepackage{hyperref}                         
\usepackage{url}                              
\usepackage[
  style=alphabetic,
  backend=biber,
  maxalphanames=3,
  maxbibnames=20,
  maxcitenames=3,
  giveninits=true,
  doi=false,
  url=true,
  backref=true,
]{biblatex}          
\usepackage{hyperref}  
\hypersetup{
  colorlinks=true, 
  linkcolor=blue,     
  filecolor=magenta,  
  urlcolor=cyan,      
  citecolor=purple    
}
\usepackage{amsfonts}                         
\usepackage{amsmath}  
 \usepackage{amsthm}
\usepackage{amssymb}                          
\usepackage{mathrsfs}                         
\usepackage{bm}                               
\usepackage{nicefrac}                         
\usepackage{xfrac}                            
\usepackage{subdepth}                         
\usepackage{array}                            
\usepackage{booktabs}                         
\usepackage{etoolbox}                         
\usepackage{microtype}                        
\usepackage[nameinlink,capitalise]{cleveref}  
\usepackage[font=small,labelfont=bf]{caption} 
\usepackage[format=hang]{subcaption}          
\usepackage{graphicx}                         
\usepackage[export]{adjustbox}                
\usepackage{color}                            
\usepackage{xcolor}                           
\usepackage{wrapfig}
\usepackage{multirow}
\usepackage{ifthen}                           
\usepackage{sidecap}
\crefname{lemma}{lemma}{lemmas}
\Crefname{lemma}{Lemma}{Lemmas}
\newcommand{\cf}{\textit{cf.}~}
\newcommand{\eg}{\textit{e.g.},~}
\newcommand{\ie}{\textit{i.e.},~}
\newcommand{\vs}{\textit{vs.}~}
\newcommand{\red}{\color{red}}
\newcommand{\blue}{\color{blue}}
\newcommand{\nb}[1]{{\sf\blue[#1]}}
\newcommand{\nbr}[1]{{\sf\red[#1]}}
\usepackage{marginnote}
\newcommand{\ignore}[1]{}

\definecolor{electric-purple}{RGB}{191, 0, 255}
\newcommand{\erin}[1]{{\sf\color{electric-purple}[Erin: #1]}}
\definecolor{kelly-green}{RGB}{45, 179, 0}

\renewcommand{\nb}[1]{}
\renewcommand{\nbr}[1]{}
\renewcommand{\erin}[1]{}
\title{
  Nonlinear dynamics of localization in \\ neural receptive fields
}
\author{
	Leon Lufkin \\
	Yale University \\
	\texttt{leon.lufkin@yale.edu} \\
    	\And
	Andrew Saxe \\
	Gatsby Unit \& SWC, UCL \\
	\texttt{a.saxe@ucl.ac.uk} \\
	\And
	Erin Grant \\
	Gatsby Unit \& SWC, UCL \\
	\texttt{erin.grant@ucl.ac.uk} 
}
\usepackage{tcolorbox}
\usepackage{enumitem}
\tcbuselibrary{theorems, breakable, skins}
\tcbset{
  assumption-box/.style={
    enhanced,
    colback=white, colframe=black,
    colbacktitle=white, coltitle=black,
    boxed title style={size=small,colframe=white},
    boxrule=0.3mm,
    rounded corners,
    before upper={\parindent0pt},
    separator sign none,
    attach boxed title to top center={yshift=-3mm},
    before=\vspace{0pt},
    after=\vspace{5pt},
  }
}
\newtcbtheorem[no counter]{model}{}{
  assumption-box,
  before upper={
    \setlist[enumerate]{
      leftmargin=*,
      label=(M\arabic*), 
      ref=M\thetcbcounter\arabic*, 
    }
  },
}{prop}
\newtcbtheorem[no counter]{stimulus}{}{
  assumption-box,
  before upper={
    \setlist[enumerate]{
      leftmargin=*,
      label=(S\arabic*), 
      ref=S\thetcbcounter\arabic*, 
    }
  },
}{prop}
\newtcbtheorem[no counter]{analysis}{}{
  assumption-box,
  before upper={
    \setlist[enumerate]{
      leftmargin=*,
      label=(A\arabic*), 
      ref=A\thetcbcounter\arabic*, 
    }
  },
}{assumption}

\begin{document}

\maketitle

\begin{abstract}
  Localized receptive fields---neurons that are selective for certain contiguous spatiotemporal features of their input---populate early sensory regions of the mammalian brain. Unsupervised learning algorithms that optimize explicit sparsity or independence criteria replicate features of these localized receptive fields, but fail to explain directly how localization arises through learning without efficient coding, as occurs in early layers of deep neural networks and might occur in early sensory regions of biological systems. We consider an alternative model in which localized receptive fields emerge without explicit top-down efficiency constraints---a feedforward neural network trained on a data model inspired by the structure of natural images. Previous work identified the importance of non-Gaussian statistics to localization in this setting but left open questions about the mechanisms driving dynamical emergence. We address these questions by deriving the effective learning dynamics for a single nonlinear neuron, making precise how higher-order statistical properties of the input data drive emergent localization, and we demonstrate that the predictions of these effective dynamics extend to the many-neuron setting. Our analysis provides an alternative explanation for the ubiquity of localization as resulting from the nonlinear dynamics of learning in neural circuits.\smash{\footnotemark}\footnotetext{
Code 
to replicate experiments and figures 
at
\url{https://github.com/leonlufkin/localization}.
}

\end{abstract}

\section{Introduction}
\label{sec:introduction}

A striking feature of peripheral responses in the animal nervous system is \emph{localization}---that is,
the linear receptive fields of simple-cell neurons often respond to contiguous regions much smaller than their full input domain. 
In vision, retinal ganglion cells approximate localized center-surround filters that tile the input~\parencite{dacey2000center,doi2012efficient,knudsen1978space},
and simple cells downstream in primary visual cortex have localized filters that are selective for spatial frequency and orientation~\parencite{hubel1959receptive,hubel1968receptive,rolls1995sparseness,niell2008highly,willmore2011sparse,ringach2002orientation,ringach2002spatial}.
In primary somatosensory cortex, neurons respond to stimulation of restricted regions of skin~\parencite{crochet2011synaptic} and
in primary auditory cortex, spatiotemporal receptive fields are typically localized in both time and frequency domains~\parencite{deweese2003binary,hromadka2008sparse};
see \cref{fig:sim-real-gabors} (left).

By contrast, artificial learning systems do not always learn localized filters. 
Principal component analysis tends to fit weights that span the entire input signal, as do unregularized autoencoder neural network architectures and restricted Boltzmann machines~\parencite{saxe2011unsupervised}.
This difference has prompted the search for artificial learning models that can learn 
localized receptive fields from naturalistic stimuli,
the most notable of which are sparse coding~\parencite{olshausen1996emergence,olshausen1997sparse}
and independent component analysis~\parencite[ICA;~][]{bell1997independent,vanhateren1998independent}.
Sparse coding, ICA, and related compression methods that produce localized receptive fields from naturalistic data share a top-down approach---they find an efficient representation of the input signal by optimizing an explicit sparsity criterion, or an independence criterion that necessitates sparsity in a critically parameterized regime~\cite{field1999wavelets,saxe2011unsupervised}.

Though sparsity is appealing as a potentially unifying explanation for localization, localization also emerges naturally in networks trained to perform classification tasks without any explicit sparsity regularization~\parencite{krizhevsky2012imagenet,zeiler2013visualizing,yosinski2015understanding,sengupta2018manifoldtiling}; see \cref{fig:sim-real-gabors} (center) for an example.
\textcite{ingrosso2022data} distilled such examples of emergent localization by demonstrating that localized receptive fields emerge in simple feedforward neural networks trained on a data model with properties meant to approximate natural visual input, in particular,
locality structure (statistical independence of non-collocated dimensions)
and non-Gaussianity (higher-order cumulants are non-null).
In simulations, \textcite{ingrosso2022data} tie the dynamical emergence of localization to increased tuning to higher-order statistics of the input, and demonstrate that even a single neuron is sufficient to learn a localized receptive field in this setting.

In this work, we build on the demonstration of \textcite{ingrosso2022data} with the aim of describing the mechanisms behind the emergence of a localized receptive field in this minimal setting.
The higher-order input statistics that drive localization are challenging to analyze with existing tools that exploit implied Gaussianity~\parencite{goldt2020modelling}.
By separating two stages of learning, we are able to derive equations for the effective early-time learning dynamics of the single neuron model that learns a localized receptive field from idealized naturalistic data.
Our analytical model identifies a concise description of the higher-order statistics that drive emergence, 
and we validate both positive and negative predictions of this analytical model via simulations with many neurons; see \cref{fig:sim-real-gabors} (right).
These findings suggest an alternative path to account for the ubiquity of localization in early neural responses as resulting from the interaction of the nonlinear dynamics of learning in neural circuits and naturalistic data with higher-order statitistical structure,
rather than an explicit efficiency criterion.

\newcommand{\fieldsize}{22pt}
\begin{figure}[t]
  \centering
  \begin{subfigure}[b]{0.15\textwidth}
    \setlength{\tabcolsep}{2pt}
    \begin{tabular}{cc}
      \tcbincludegraphics[size=tight,hbox,graphics options={width=\fieldsize}]{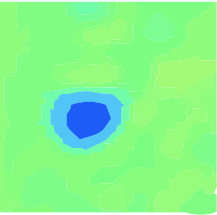} &
      \tcbincludegraphics[size=tight,hbox,graphics options={width=\fieldsize}]{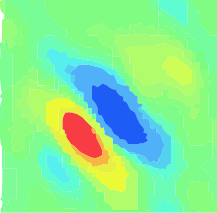} \\
      \tcbincludegraphics[size=tight,hbox,graphics options={width=\fieldsize}]{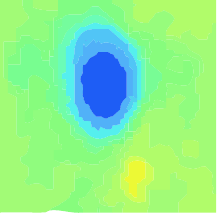} &
      \tcbincludegraphics[size=tight,hbox,graphics options={width=\fieldsize}]{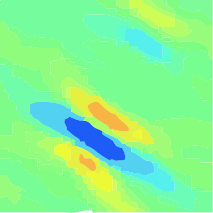} \\
      \tcbincludegraphics[size=tight,hbox,graphics options={width=\fieldsize}]{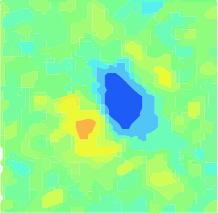} &
      \tcbincludegraphics[size=tight,hbox,graphics options={width=\fieldsize}]{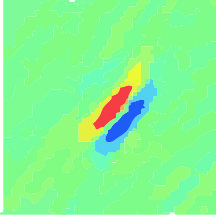} \\
      \tcbincludegraphics[size=tight,hbox,graphics options={width=\fieldsize}]{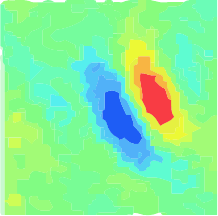} &
      \tcbincludegraphics[size=tight,hbox,graphics options={width=\fieldsize}]{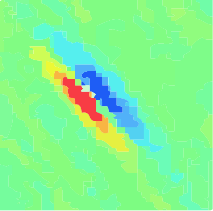} \\
      \multicolumn{2}{c}{\scriptsize\parencite{ringach2002spatial}}
    \end{tabular}
  \end{subfigure}
  \begin{subfigure}[b]{0.09\textwidth}
    \setlength{\tabcolsep}{2pt}
    \begin{tabular}{cc}
      \tcbincludegraphics[size=tight,hbox,graphics options={width=\fieldsize}]{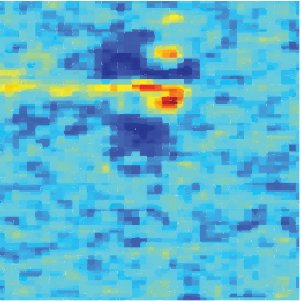} \\
      \tcbincludegraphics[size=tight,hbox,graphics options={width=\fieldsize}]{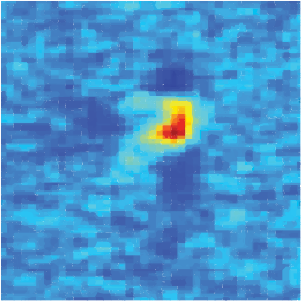} \\
      \tcbincludegraphics[size=tight,hbox,graphics options={width=\fieldsize}]{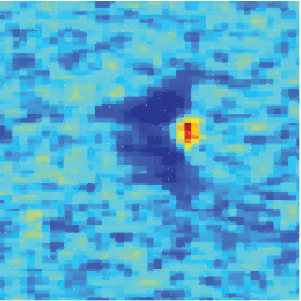} \\
      \tcbincludegraphics[size=tight,hbox,graphics options={width=\fieldsize}]{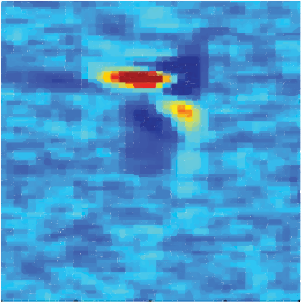} \\
      \scriptsize\parencite{decharms1998optimizing}
    \end{tabular}
  \end{subfigure}
  \begin{subfigure}[b]{0.09\textwidth}
    \setlength{\tabcolsep}{2pt}
    \begin{tabular}{cc}
      \tcbincludegraphics[size=tight,hbox,graphics options={width=\fieldsize}]{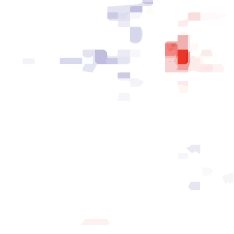} \\
      \tcbincludegraphics[size=tight,hbox,graphics options={width=\fieldsize}]{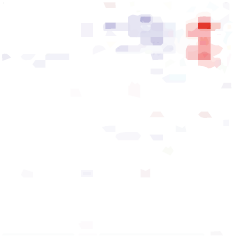} \\
      \tcbincludegraphics[size=tight,hbox,graphics options={width=\fieldsize}]{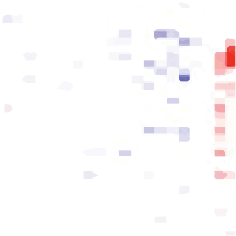} \\
      \tcbincludegraphics[size=tight,hbox,graphics options={width=\fieldsize}]{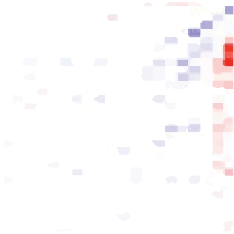} \\
      \scriptsize\parencite{singer2018sensory}
    \end{tabular}
  \end{subfigure}
  \begin{subfigure}[b]{0.03\textwidth}
    \raisebox{-\height/2 + 6pt}{
    \begin{tikzpicture}
      \draw[dashed, gray, ultra thick] (0,0) -- (0,4);
    \end{tikzpicture}
    }
  \end{subfigure}
  \begin{subfigure}[b]{0.24\textwidth}
    \begin{tabular}{c}
      \includegraphics[width=79pt]{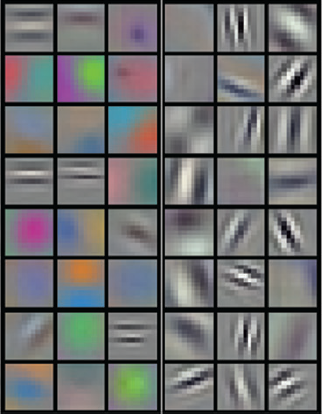} \\
      \scriptsize\parencite{krizhevsky2012imagenet}
    \end{tabular}
  \end{subfigure}
  \begin{subfigure}[b]{0.04\textwidth}
    \raisebox{-\height/2 + 6pt}{
  \begin{tikzpicture}
    \draw[dashed, gray, ultra thick] (0,0) -- (0,4);
  \end{tikzpicture}
}
  \end{subfigure}
  \begin{subfigure}[b]{0.16\textwidth}
    \setlength{\tabcolsep}{2pt}
    \begin{tabular}{cc}
      \tcbincludegraphics[size=tight,hbox,graphics options={width=\fieldsize}]{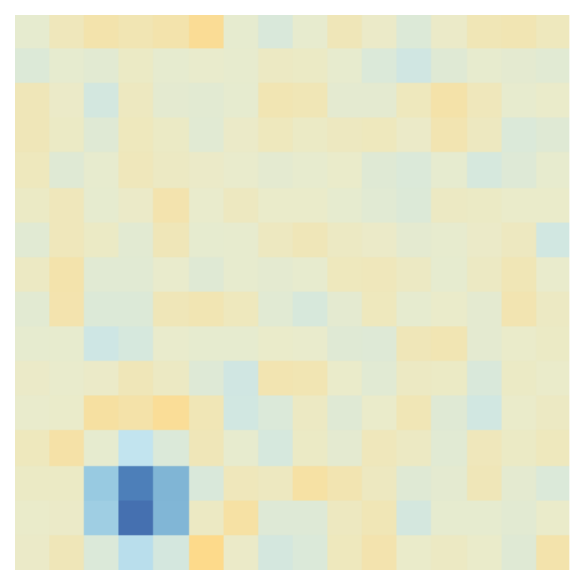} &
      \tcbincludegraphics[size=tight,hbox,graphics options={width=\fieldsize}]{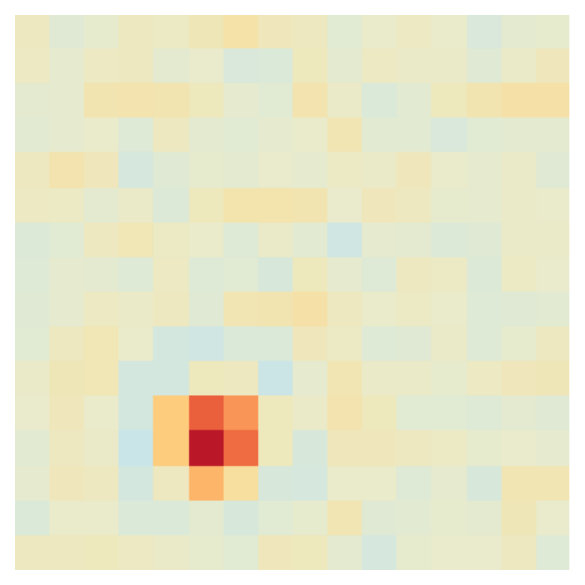} \\
      \tcbincludegraphics[size=tight,hbox,graphics options={width=\fieldsize}]{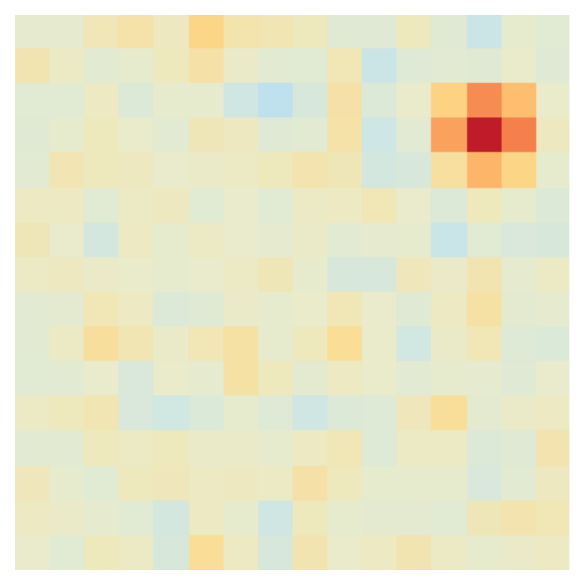} &
      \tcbincludegraphics[size=tight,hbox,graphics options={width=\fieldsize}]{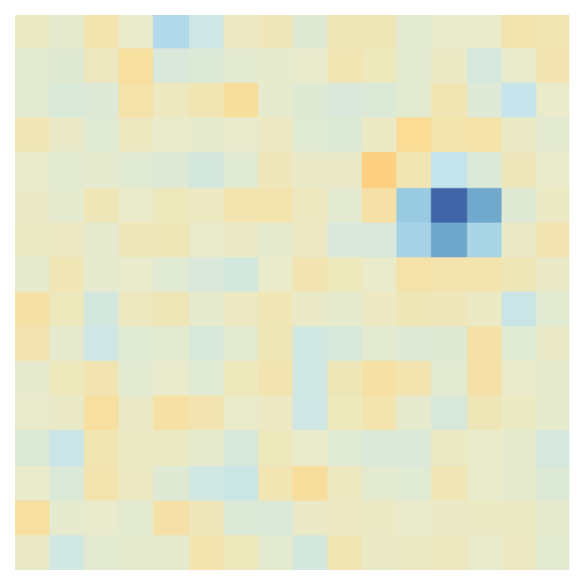} \\
      \tcbincludegraphics[size=tight,hbox,graphics options={width=\fieldsize}]{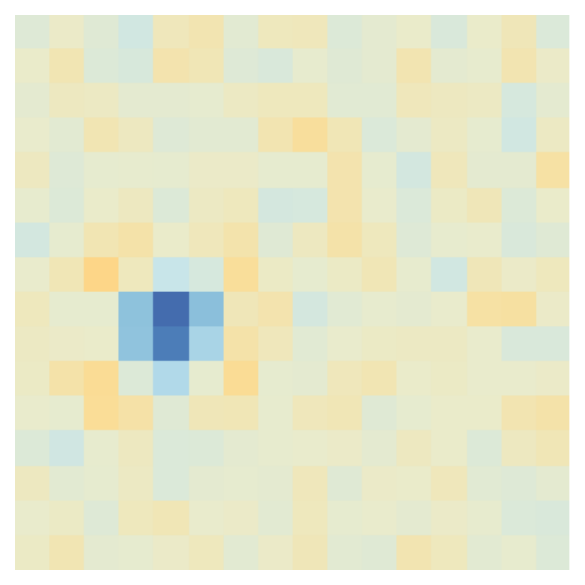} &
      \tcbincludegraphics[size=tight,hbox,graphics options={width=\fieldsize}]{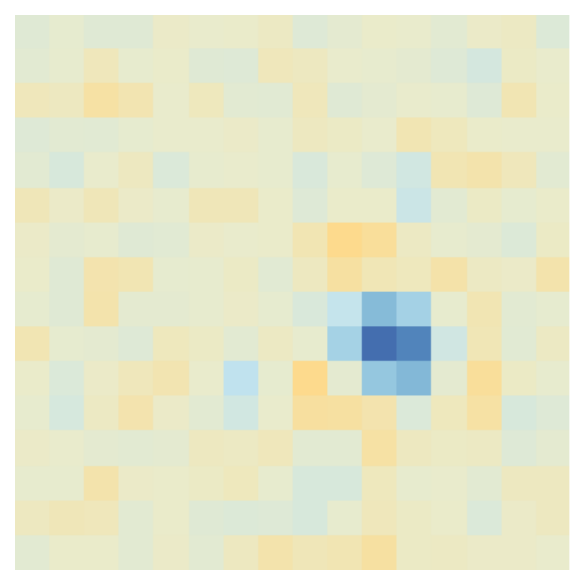} \\
      \tcbincludegraphics[size=tight,hbox,graphics options={width=\fieldsize}]{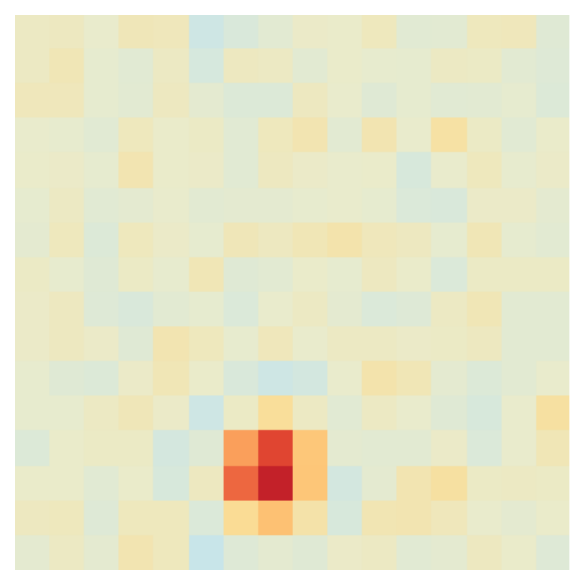} &
      \tcbincludegraphics[size=tight,hbox,graphics options={width=\fieldsize}]{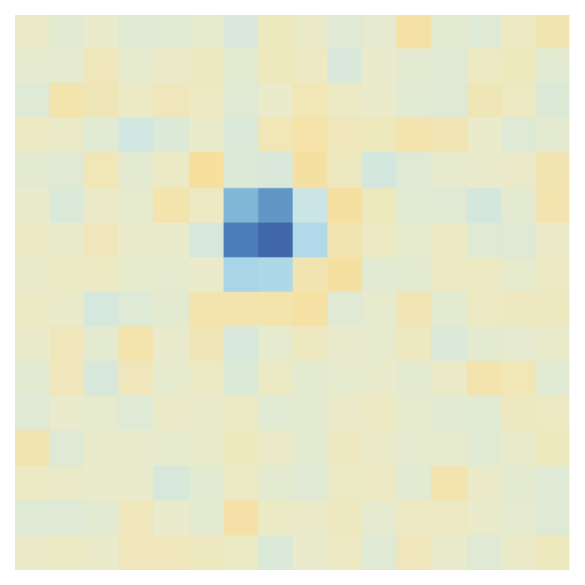} \\
      \multicolumn{2}{c}{\scriptsize\parencite{hyvarinen2000independent}}
    \end{tabular}
  \end{subfigure}
  \begin{subfigure}[b]{0.15\textwidth}
    \setlength{\tabcolsep}{2pt}
    \begin{tabular}{cc}
      \tcbincludegraphics[size=tight,hbox,graphics options={width=\fieldsize}]{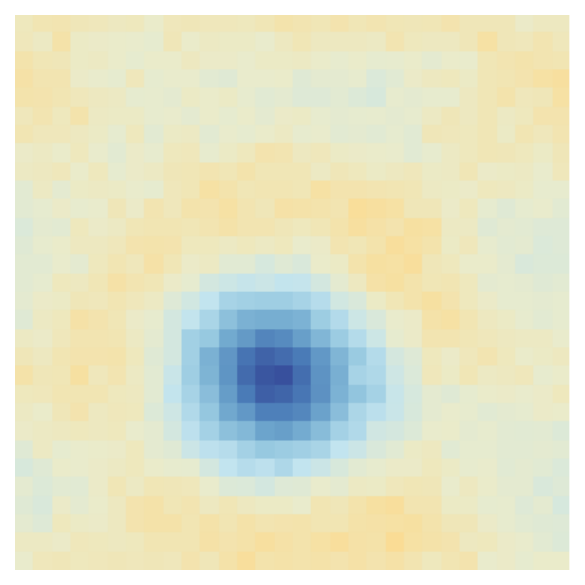} &
      \tcbincludegraphics[size=tight,hbox,graphics options={width=\fieldsize}]{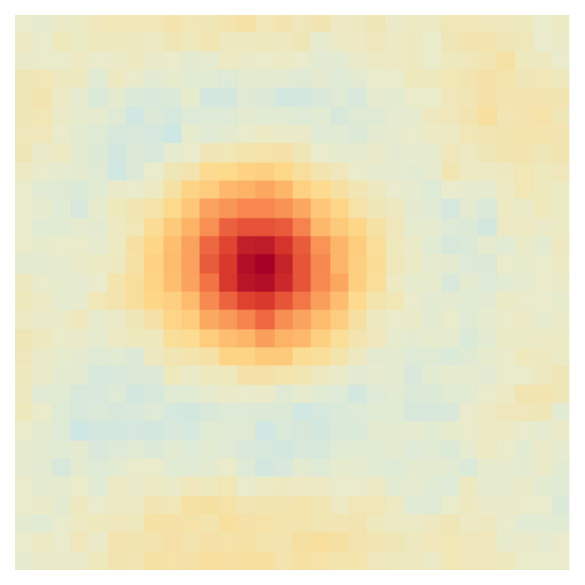} \\
      \tcbincludegraphics[size=tight,hbox,graphics options={width=\fieldsize}]{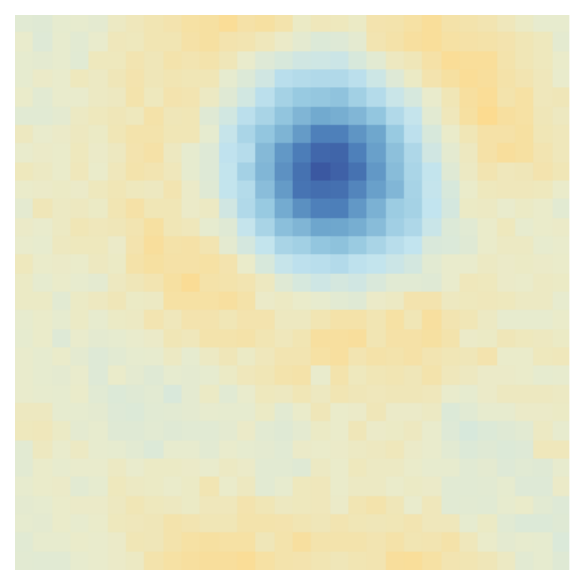} &
      \tcbincludegraphics[size=tight,hbox,graphics options={width=\fieldsize}]{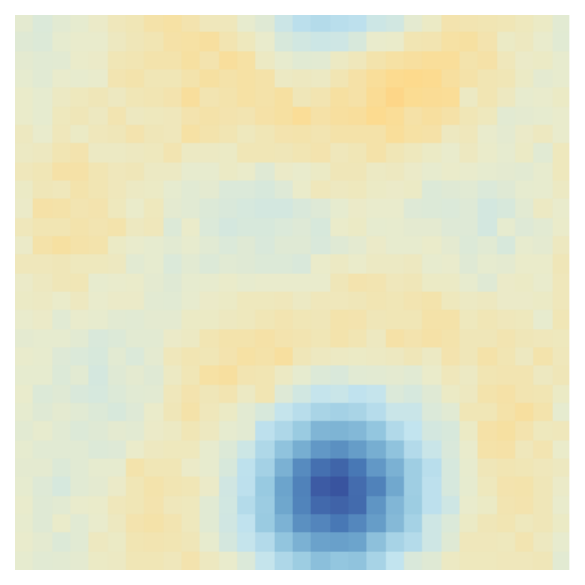} \\
      \tcbincludegraphics[size=tight,hbox,graphics options={width=\fieldsize}]{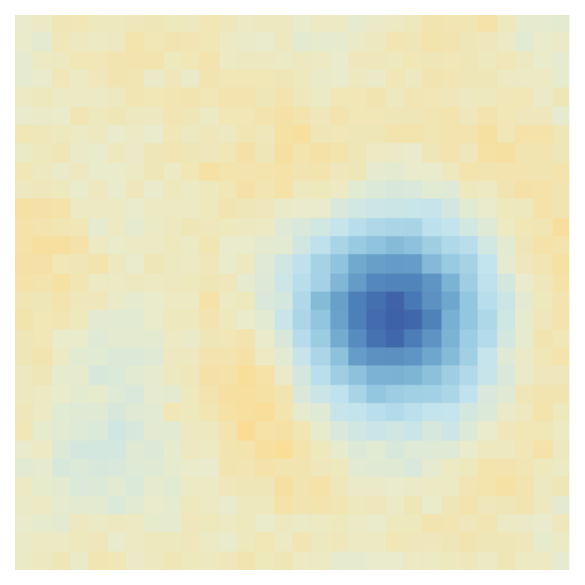} &
      \tcbincludegraphics[size=tight,hbox,graphics options={width=\fieldsize}]{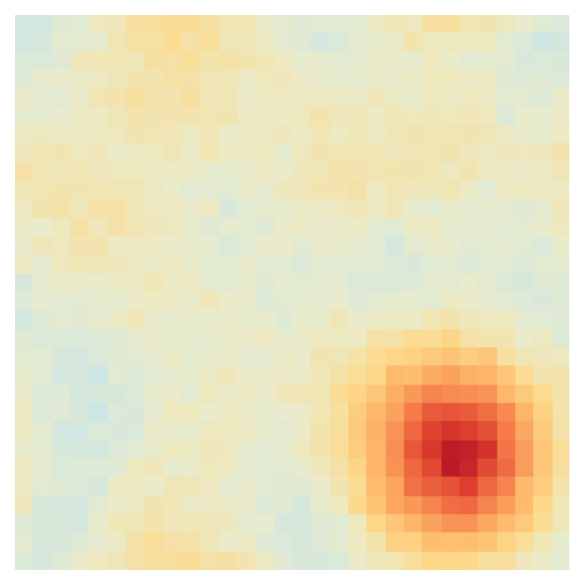} \\
      \tcbincludegraphics[size=tight,hbox,graphics options={width=\fieldsize}]{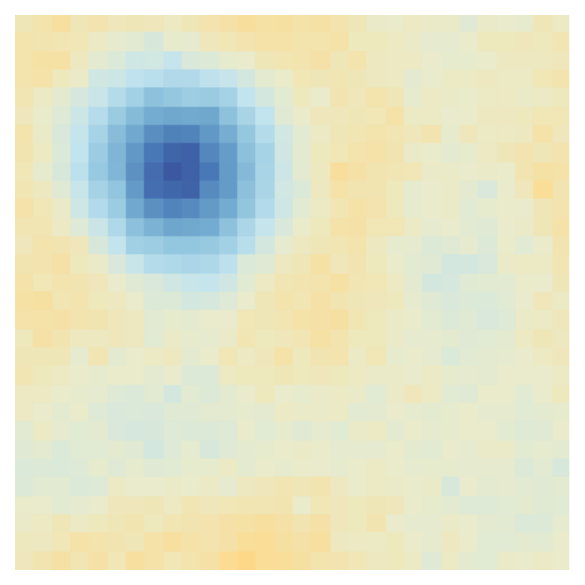} &
      \tcbincludegraphics[size=tight,hbox,graphics options={width=\fieldsize}]{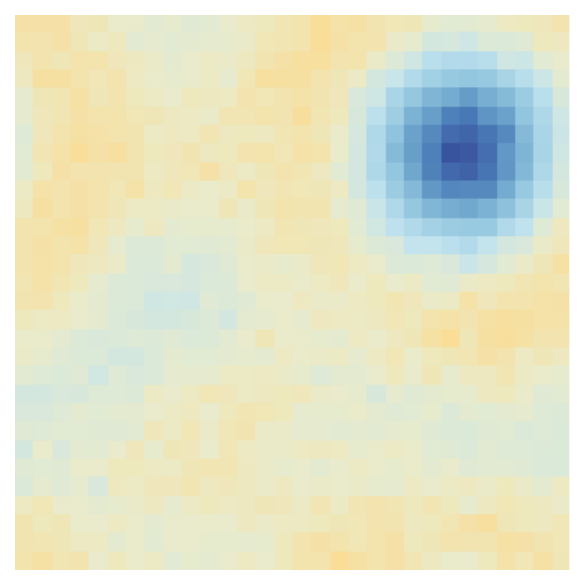} \\
      \multicolumn{2}{c}{\scriptsize\labelcref{item:many-neuron-model} SCM}
    \end{tabular}
  \end{subfigure}
  \caption{
    \textbf{(Left)}
    Localization in
    spatial receptive fields (RFs) measured from non-human primate (NHP) primary visual cortex~\parencite[][Fig.~2]{ringach2002spatial}
    and in spatiotemporal RFs measured from NHP~\parencite[][Fig.~2]{decharms1998optimizing} and ferret~\parencite[][Fig.~2]{singer2018sensory} primary auditory cortex.
    \textbf{(Center)}
    Half-slice of the localized first-layer kernels of AlexNet trained for ImageNet classification~\parencite{krizhevsky2012imagenet}.
    \textbf{(Right)}
    Localized receptive fields learned from the task of \cref{sec:task} in 2-D
    using ICA~\parencite{hyvarinen2000independent} 
    and the soft committee machine (SCM; \labelcref{item:many-neuron-model} with fixed second-layer weights)
    of \cref{sec:model}.
    \emph{Localization---spatial and/or temporal selectivity---appears across settings,
      as measured by response maximization in biological systems (left) and by inspecting linear filters in artificial systems (center, right).}
  }
  \label{fig:sim-real-gabors}
  \vspace{-8pt}
\end{figure}

\section{Modeling approach}
\label{sec:prelims}

We extend the setting of \textcite{ingrosso2022data},
a minimal example of a neural network that learns localized receptive fields 
from idealized naturalistic data.
We analyze the dynamics of learning in this setting in \cref{sec:theory}
and validate our analytical model with simulations in \cref{sec:experiments}.

\subsection{Neural network architecture and learning algorithm}
\label{sec:model}

We consider a two-layer feedforward neural network with nonlinear activation and scalar output.
While simple, this architecture is highly expressive, capable of approximating arbitrary 
integrable univariate functions with appropriate scaling~\parencite{barron1993universal, pinkus1999approximation},
and exhibits rich feature learning dynamics that underlie the performance of models at scale~\parencite{woodworth2020kernel},
making this architecture the ongoing subject of theoretical neural network analyses~\parencite{mei2018mean, goldt2019dynamics, veiga2022phase}.
We denote a two-layer network with $N$-dimensional input, $M$ hidden units, and one-dimensional scalar output as
\newcounter{modelenumi}
\begin{model}{\textbf{Model 1} (\emph{many-neuron architecture}).}{}
\begin{enumerate}[series=modelenumi]
  \item \label{item:many-neuron-model} 
    \makebox[\linewidth - 2.5em]{
      $\hat{y}(\mathbf{x}) = b^{(2)} + \sum_{m=1}^M w_m^{(2)} 
      \sigma\left(b_m^{(1)} + \langle \mathbf{w}_m^{(1)}, \mathbf{x} \rangle \right)$
    }
\end{enumerate}
\end{model}
where $\sigma : \R \to \R$ is a pointwise nonlinearity such as the rectified linear unit (ReLU) or sigmoid function,
$\mathbf{w}_m^{(1)} \in \R^N$ and $w_m^{(2)} \in \R$ are learnable weights, 
$b_m^{(1)}, b^{(2)} \in \R$ are learnable bias terms, and
$\langle \cdot, \cdot \rangle$ denotes the standard Euclidean inner (dot) product on $\R^N$.
When the second-layer parameters are fixed, this model is known as a \emph{soft-committee machine}~\parencite[SCM;][]{saad1995line},
which~\parencite{ingrosso2022data} notes learns less noisy receptive fields but exhibits similar localization behavior.
The many-neuron architecture in \labelcref{item:many-neuron-model} is the focus of our \textbf{simulations}~(\cref{sec:experiments}), but the dynamics of this model 
are too complex to analyze directly, even for the idealized naturalistic data model considered here.
In order to derive \textbf{analytical} results~(\cref{sec:theory}), we consider the simplest neural network exhibiting the desired localization phenomenon, a single hidden neuron without bias and with rectified linear unit activation, written as
\begin{model}{\textbf{Model 2} (\emph{single-neuron architecture}).}{}
\begin{enumerate}[resume*=modelenumi]
  \item \label{item:single-neuron-model} 
    \makebox[\linewidth - 2.5em]{
      $\hat{y}(\mathbf{x}) = 
      \operatorname{ReLU}\left(\langle \mathbf{w}, \mathbf{x} \rangle \right)$
    }
\end{enumerate}
\end{model}
where $\operatorname{ReLU}(x) = \max(x,0)$, applied pointwise to vectorial input.
As \textcite{ingrosso2022data} demonstrate, the localized receptive fields learned by the 
many- and single-neuron models defined in \labelcref{item:many-neuron-model,item:single-neuron-model} are qualitatively similar up to spatial translation, 
which permits us to generalize insights from analyzing the learning dynamics of the single-neuron~\labelcref{item:single-neuron-model} to the many-neuron~\labelcref{item:many-neuron-model}.
For simulations, we initialize the weights and biases
as independent draws from an isotropic Gaussian distribution with scaled variance,
and train with batch gradient descent with a fixed learning rate on the mean-squared error
(MSE) evaluated on input-output pairs from the task; see \cref{sec:task} for task sampling procedures.

\subsection{Stimulus properties}
\label{sec:input}

The data model of \textcite{ingrosso2022data} 
can be shown to satisfy three conditions that enable the analysis we give in \cref{sec:theory}.
We consider several other data models that share the below properties but differ in generative mechanism
in order to probe the effect of these properties on localization.
In particular, we consider data $\mathbf{X}$ sampled from distributions $p$ on $\R^N$ satisfying the following:
\newcounter{propenumi}
\begin{stimulus}{\textbf{Stimulus properties 1--3} (\emph{idealization of natural images}).}{}
\begin{enumerate}[series=propenumi]
  \item \label{item:weak-dependence} (Positional) weak dependence: for any fixed $\rho \in (0,1)$, as $N \to \infty$,
    $$\alpha(N) \triangleq \sup_{A \subseteq \R, B \subseteq \R^{(1-\rho) N}} |\PR(X_1 \in A, X_{> \rho N} \in B) - \PR(X_1 \in A) \PR(X_{> \rho N} \in B)| \to 0~,$$
  \item \label{item:translation-invariance} Translation invariance: $p(\mathbf{X} = \mathbf{x}) = p(\mathbf{X} = \mathcal{S} \mathbf{x})$ for all $\mathbf{x} \in \R^N$, where $\mathcal{S}$ is the circular shift operator, and
  \item \label{item:sign-symmetry} Sign symmetry: $p(\mathbf{X} = \mathbf{x}) = p(\mathbf{X} = -\mathbf{x})$ for all $\mathbf{x} \in \R^N$.
\end{enumerate}
\end{stimulus}
Properties~\labelcref{item:weak-dependence,item:translation-invariance}
are defining characteristics of natural image data~\parencite{hyvarinen2009natural}.
Property~\labelcref{item:sign-symmetry}
can also be seen to hold for natural images after centering
and is convenient analytically
because it implies that $\E[\mathbf{X}] = 0$.
Property~\labelcref{item:weak-dependence} assumes that $p$ is implicitly parameterized by $N$
in order to state that the statistical dependence between entries of $\mathbf{X}$ vanishes 
as their separation increases.\smash{\footnotemark}\footnotetext{
  The weak dependence condition in \labelcref{item:weak-dependence} is based on strong $\alpha$-mixing, a notion first introduced by \cite{rosenblatt1956central} 
  to obtain a generalization of the central limit theorem, which we employ later on.
  We choose $\alpha$-mixing because it is easy to interpret and verify,
  but alternative definitions of weak dependence \parencite[\eg][]{bardet2008dependent} can be substituted.
}

We denote the covariance of $\mathbf{X}$ by $\Sigma \triangleq \operatorname{Cov}[\mathbf{X}]$,
the square of principal-diagonal entries (the variance of each entry of $\mathbf{X}$) by $\sigma^2$, 
and the $i$-th row by $\sigma_i$.
Weak dependence (\labelcref{item:weak-dependence}) implies that entries far from the principal diagonal of $\Sigma$ will be 0, while translation-invariance (\labelcref{item:translation-invariance}) implies that $\Sigma$ is circulant (\ie entries along each diagonal are equal) and thus identifiable by a single row; see \cref{fig:task} (center).

\subsection{Lengthscale discrimination task}
\label{sec:task}

\textcite{ingrosso2022data} develop a minimal task for which localization emerges in a feedforward neural network: binary discrimination between inputs from two distributions that differ in the lengthscale of the correlations between their entries.
This lengthscale discrimination task can be seen as a pretext task for self-supervised learning~\parencite{kolesnikov2019revisiting,chen2020simple} of representations~\parencite[\cf~unsupervised:][]{olshausen1996emergence,bell1997independent}.
More precisely, we generate data $(\mathbf{X},Y)$ for supervised training according to
\begin{align} \label{eq:task}
    \mathbf{X} \mid Y = y \sim p(\mathbf{X};\Sigma_y)~,
\end{align}
where $p$ is to be defined, $\Sigma_y$ are distinct covariance matrices for each $y$, and we sample $Y$ uniformly among a set of increasing \emph{lengthscale correlation classes} $y \in \{0,1,\ldots\}$, which correspond to the strength of correlation between distant positions.
For instance, in the case of two classes ($y = 0, 1$), we take $\Sigma_0$ to be closer to $\sigma^2 \mathbb{I}_N$ than $\Sigma_1$, where $\mathbb{I}_N$ is the $N \times N$ identity matrix and $\sigma$ is a fixed value.
This construction isolates, via distinct covariance matrices per class, the second-order statistics, which we will see below enter into the learning dynamics separately from other properties of $p(\mathbf{X})$, including, most critically, the implied marginal distributions, $p(X_i)$.

\paragraph{\texttt{Ising}.}\hspace{-2pt}
The first distribution we consider is the one-dimensional Ising model.
It is of interest as a distribution that satisfies \labelcref{item:weak-dependence,item:translation-invariance,item:sign-symmetry} with marginals $p(X_i)$ with extreme support on $\{ \pm 1 \}$,
making it the distribution that promotes localization most strongly, as we will see in \cref{sec:theory}.
In the absence of an external field, the Ising distribution is
\begin{equation}
    p_\text{\texttt{Ising}}(\mathbf{X}=\mathbf{x})
    = p_\text{\texttt{Ising}}(X_1=x_1,\ldots,X_N=x_N) = e^{ -\sum_{i=1}^{N} J x_i x_{i+1} } / \mathcal{Z},
\end{equation}
where $J$ is a chosen pairwise interaction strength, $\mathcal{Z}$ is the normalizing constant, and we enforce a periodic boundary constraint via $x_{N+1} \equiv x_1$.
As $J$ increases, the lengthscale of the correlations in $\mathbf{X}$ also increases.
For simulations, we sample from $p_\texttt{Ising}$ using a Gibbs sampler~\parencite{geman1984stochastic}.
Discrimination tasks in the simulations in 
\cref{sec:experiments}
use $J_1=0.7$ (for $y=1$) and $J_0=0.3$ (for $y=0$).

\newcommand{\sampleheight}{42pt}
\newcommand{\covheight}{46pt}
\newcommand{\marginalheight}{50pt}
\setlength{\tabcolsep}{4pt}
\begin{figure}[t]
  \centering
  \hspace{-1.2em}
  \scalebox{0.9}{
  \begin{centering}
    \begin{tabular}{p{51pt}
      @{\hspace{10pt}}m{2pt}l
      @{\hspace{5pt}}l
      @{\hspace{10pt}}m{2pt}l
      @{\hspace{10pt}}m{2pt}l}
        \raisebox{18pt}{\small$\texttt{Ising}$} &
        \raisebox{34pt}{\rotatebox{90}{\tiny input value}} &
        \includegraphics[height=\sampleheight]{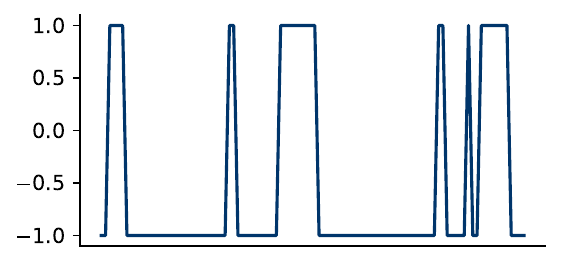} &
        \includegraphics[height=\sampleheight]{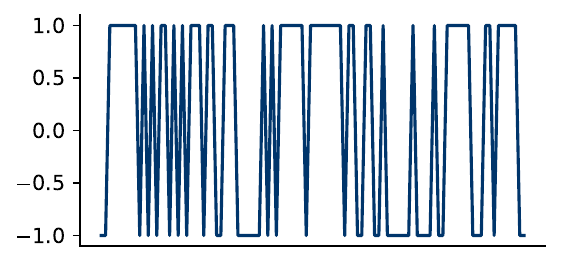} &
        \raisebox{38pt}{\rotatebox{90}{\tiny input dimension}} &
        \includegraphics[height=\covheight]{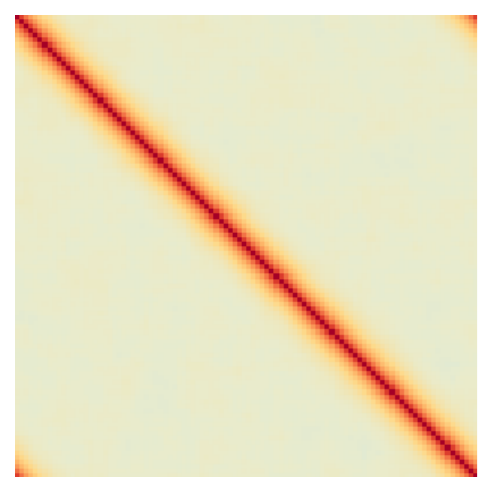} &
        \raisebox{40pt}{\rotatebox{90}{\tiny $p(X_i)$}} &
        \raisebox{-4pt}{\includegraphics[height=\marginalheight]{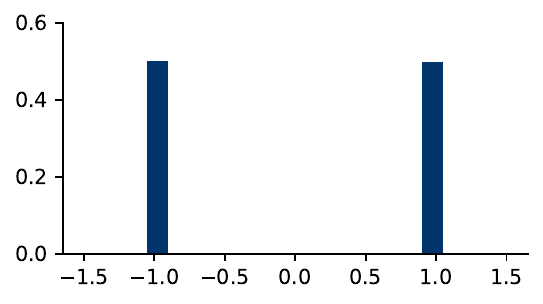}} \\
        \noalign{\vskip -36pt}
        \raisebox{18pt}{\small$\texttt{NLGP}(0.01)$} &
        \raisebox{34pt}{\rotatebox{90}{\tiny input value}} &
        \includegraphics[height=\sampleheight]{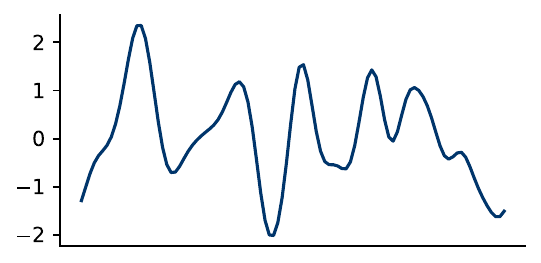} &
        \includegraphics[height=\sampleheight]{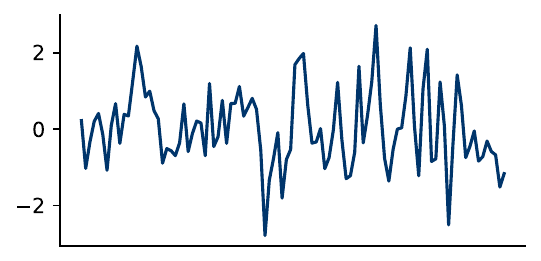} &
        \raisebox{38pt}{\rotatebox{90}{\tiny input dimension}} &
        \includegraphics[height=\covheight]{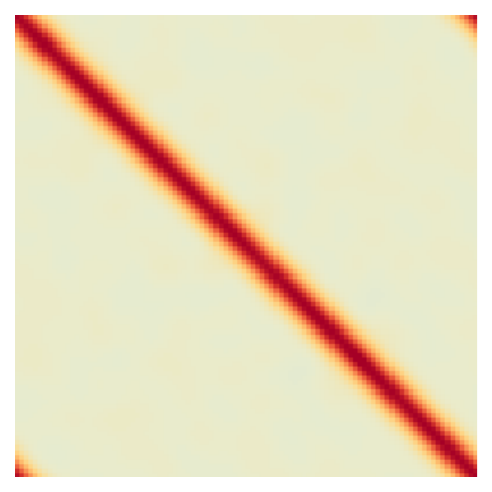} & 
        \raisebox{40pt}{\rotatebox{90}{\tiny $p(X_i)$}} &
        \raisebox{-4pt}{\includegraphics[height=\marginalheight]{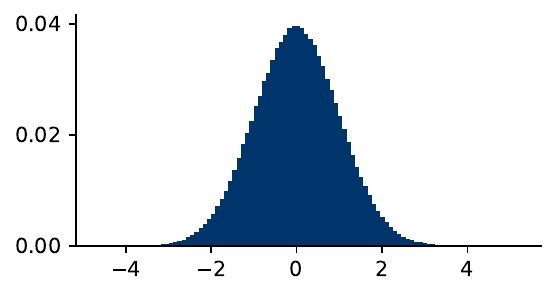}} \\
        \noalign{\vskip -36pt}
        \raisebox{18pt}{\small $\texttt{Kur}(5)$} &
        \raisebox{34pt}{\rotatebox{90}{\tiny input value}} &
        \includegraphics[height=\sampleheight]{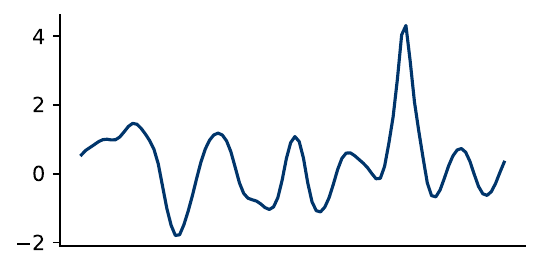} &
        \includegraphics[height=\sampleheight]{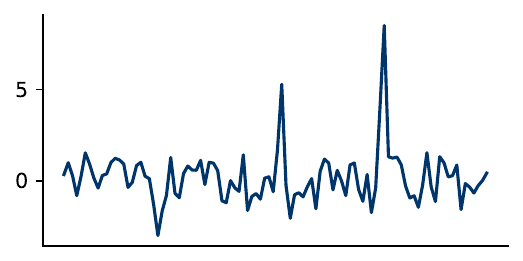} &
        \raisebox{38pt}{\rotatebox{90}{\tiny input dimension}} &
        \includegraphics[height=\covheight]{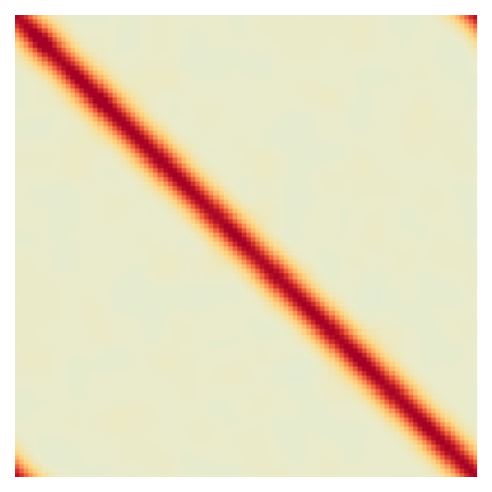} &
        \raisebox{40pt}{\rotatebox{90}{\tiny $p(X_i)$}} &
        \raisebox{-4pt}{\includegraphics[height=\marginalheight]{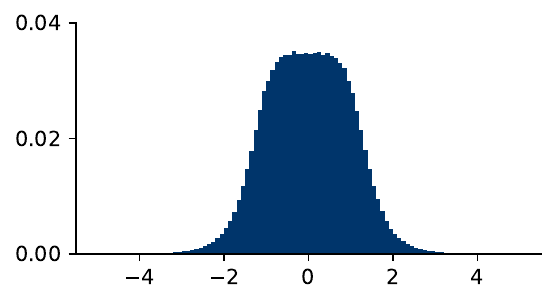}} \\
        \noalign{\vskip -37pt}
        &&
        \hspace{25pt}\tiny input dimension &
        \hspace{25pt}\tiny input dimension & &
        \hspace{3pt}\tiny input dimension & &
        \hspace{37pt}\tiny input value \\
  \end{tabular}
  \end{centering}
  }
  \caption{
    From left:
    Long- and short-lengthscale samples $\mathbf{x}$,
    covariances $\Sigma$ for one lengthscale,
    and marginals $p(X_i)$ 
    for the data models described in \cref{sec:task}:
    Ising (with $J=1.2, 0.3$ for left, right samples), 
    the nonlinear Gaussian process~\parencite[NLGP;~][]{ingrosso2022data},
    and the controllable kurtosis model, \texttt{Kur}
    (with $\xi=5, 1$ for left, right samples).
    \emph{
    Each model generates samples centered about zero and with covariances that can be constrained to be similar,
    but with differing higher-order statistics, as can be seen from the dimension-wise marginals.
    }
  }
  \label{fig:task}
  \vspace{-10pt}
\end{figure}

\paragraph{\texttt{NLGP}$(g)$.}\hspace{-2pt}
We also consider the data model used in \textcite{ingrosso2022data}, the nonlinear Gaussian process (NLGP), which enables one to interpolate between distributions that do and do not yield localization via a single parameter, $g$.
A sample $\mathbf{X} \mid Y = y$ from the NLGP is constructed by first sampling a Gaussian $\mathbf{Z} \mid Y = y \sim \NN(0, \tilde{\Sigma}_y)$ and then transforming it via
\begin{equation} \label{eq:nlgp}
    X_i \triangleq \operatorname{erf}(g Z_i) / \mathcal{Z}(g) \qquad 1 \leq i \leq N,
\end{equation}
where $\operatorname{erf}$ is the Gauss error function, $\mathcal{Z}$ is a normalization constant to ensure that the variances of $X_i$ and $Z_i$ are the same, and $\tilde{\Sigma}_y$ is a covariance matrix for $\mathbf{Z}$, where we use $(\tilde{\Sigma}_y)_{ij} = \exp(-(i-j)^2/\xi^2)$ for a lengthscale parameter $\xi$ \cite{ingrosso2022data}. 
If $g \approx 0$ (where localization is \emph{not} observed), $g Z_i$ will tend to lie in the linear regime of $\operatorname{erf}$, so $Z_i$ will be untransformed, \ie $\mathbf{X}$ is Gaussian.
However, as $g \to \infty$ (where localization \emph{is} observed), $g Z_i$ will tend to saturate $\operatorname{erf}$, so $X_i$ will have support on $\{ \pm 1 \}$.

\paragraph{\texttt{Kur}$(k)$.}\hspace{-2pt}
The final family we consider is chosen to give us flexibility over the kurtosis $\kappa$ of the marginals $p(X_i)$.
In the Ising model, the \emph{excess} kurtosis ($\kappa - 3$) of the marginals is fixed at $-2$, while in $\texttt{NLGP}(g)$, it varies from $-2$ to $0$.
This family allows us to vary the excess kurtosis from negative through positive values.
We sample $\mathbf{X} \mid Y = y$ from this family via inverse transform sampling to vary the marginals while enforcing dependence via Gaussian copulas.
More concretely, we sample $\mathbf{Z} \mid Y = y \sim \NN(0, \tilde{\Sigma}_y)$ and then transform it via
\begin{equation}
    X_i \triangleq f^{-1}( \Phi( Z_i / \tilde{\sigma} )) / \mathcal{Z}, \qquad 1 \leq i \leq N, \label{eq:alg}
\end{equation}
where $\tilde{\sigma}$ is the standard deviation of $Z_i$, $\Phi$ is the standard Gaussian cumulative distribution function (CDF), $f$ is the CDF of the desired marginal distribution for $X_i$, and $\mathcal{Z}$ is a normalization constant, which we compute numerically.
We define $\tilde{\Sigma}_y$ as for $\texttt{NLGP}$.
We choose $f$ to be the generalized \emph{algebraic sigmoid} function (see \cref{sec:algebraic-sigmoid})
for $k > 0$ to make use of its tractable inverse, simplifying the procedure in \cref{eq:alg}.
We denote the corresponding distribution by $\texttt{Kur}(k)$.
Though we are able to continuously vary excess kurtosis, we lack an explicit form; however, numerical computation shows that for $k \lessapprox 5.8$, excess kurtosis is positive, while for $k \gtrapprox 5.9$, it is negative.

\section{Theoretical results}
\label{sec:theory}

We derive an analytical model for the localization dynamics of the single-neuron architecture in \labelcref{item:single-neuron-model}.
This result establishes necessary and sufficient conditions for localization under Properties~\labelcref{item:weak-dependence,item:translation-invariance,item:sign-symmetry} for the minimal case of a binary response, \ie $y = 0,1$.
The conditions for localization in the single-neuron architecture in \labelcref{item:single-neuron-model} are demonstrated in \cref{sec:experiments} to also hold empirically for the many-neuron architecture in \labelcref{item:many-neuron-model}.
Further, we use this model to derive a negative prediction about localization, that the architectures in \labelcref{item:many-neuron-model,item:single-neuron-model} fail to learn a localized receptive field on elliptical distributions
despite their non-Gaussian---in particular, significantly positive kurtosis---statistics~\parencite[\cf positive kurtosis as an objective or diagnostic for localization,][]{hyvarinen2000independent,ingrosso2022data}.

\subsection{An analytical model for the dynamics of localization in a single neuron}

Previous approaches to obtain analytical dynamics in the architectures in 
\labelcref{item:many-neuron-model,item:single-neuron-model} 
have studied the gradient flow under the assumption that the preactivation $\langle \mathbf{w}, \mathbf{X} \rangle$ is approximately Gaussian~\parencite{goldt2020modelling,gerace2020generalisation,goldt2022gaussian}, but this assumption fails to capture the propagation of higher-order statistics through a neural network that promotes localization~\cite{ingrosso2022data}.
Happily, the idealized visual input setting set out in \labelcref{item:weak-dependence,item:translation-invariance,item:sign-symmetry} permits us some simplification.
In particular, 
the translation-invariance of the data $\mathbf{X}$ under \labelcref{item:translation-invariance} and 
the architecture of \labelcref{item:single-neuron-model}
allow us to work with the marginal distributions of each input dimension, $X_i$ rather than the full joint distribution of $\mathbf{X}$.

We now give the analytical simplifications that allow us to derive an analytical model for the localization dynamics of the single neuron architecture in \labelcref{item:single-neuron-model}, namely
two assumptions on $\mathbf{X} \mid X_i$ for all $i \in \{1, \dots, N\}$ as a well as a mild condition on the weights that is satisfied at initialization.
These are, where $\sigma_i^y$ to denotes the $i$-th row of $\Sigma_y$:

\newcounter{assumenumi}
\begin{analysis}{\textbf{Analytical simplifications 1--3} (\emph{early-time, limiting dynamics}).}{}
\begin{enumerate}[series=assumenumi]
  \item \label{item:mean-assumption} $\E[\mathbf{X} \mid X_i = x_i, Y = y] = x_i \sigma_i^y$, \ie the conditional mean scales linearly with $x_i$.
  \item \label{item:covariance-assumption} $\text{Cov}[\mathbf{X} \mid X_i = x_i, Y = y] = \Sigma_y - \sigma_i^y \sigma_i^{y\top}$, \ie the conditional covariance is smaller near $i$, but independent of the exact value of $x_i$.
  \item \label{item:lindeberg-condition} Lindeberg's condition holds for the sequence $w_1 X_1, \ldots, w_N X_N \mid X_i = x_i$ as $N \to \infty$ for all $x_i$.
\end{enumerate}
\end{analysis}

Our motivation for Assumptions~\labelcref{item:mean-assumption,item:covariance-assumption,item:lindeberg-condition} is that they replicate the kurtosis of the marginal distributions $X_i$ (discussed further below) of two important and distinct limiting cases where localization does and does not appear, respectively:
when $\mathbf{X}$ has support on the vertices of the hypercube $\{ \pm 1 \}^N$ (satisfied by \texttt{Ising} for any $J$), and
when $\mathbf{X}$ is Gaussian (satisfied by \texttt{NLGP} with $g \approx 0$).

The gradient flow in Lemma~\labelcref{lem:gradient_flow} also relies on Assumption~\labelcref{item:lindeberg-condition} that Lindeberg's condition holds for the sequence $w_i X_i$, which ensures that no single term $w_i X_i$ in the sequence can dominate.
If this holds, then we can conclude that $\langle \mathbf{w}, \mathbf{X} \rangle \mid X_i$ is approximately Gaussian.
As we discuss in \cref{subsec:pf_of_gradient_flow}, this is almost always satisfied for a Gaussian initialization of $\mathbf{w}$, and for slight deviations therefrom, and is satisfied by the settings of \textcite{ingrosso2022data}.
Using this fact, we obtain an explicit form for the gradient flow early in training, stated in Lemma~\labelcref{lem:gradient_flow}.

\begin{lemma} \label{lem:gradient_flow}
    Under Assumptions~\labelcref{item:mean-assumption,item:covariance-assumption},
    the gradient flow for the single ReLU neuron in \labelcref{item:single-neuron-model} early in training with $y = 0, 1$ 
    trained using MSE loss is
    \begin{equation} \label{eq:gradient_flow_early}
      \frac{2}{\tau} \frac{\mathrm{d}\mathbf{w}}{\mathrm{d}t} = \varphi\left( \frac{\Sigma_1 \mathbf{w}}{\sqrt{\langle \Sigma_1 \mathbf{w}, \mathbf{w} \rangle}} \right) - ( \Sigma_0 + \Sigma_1 ) \mathbf{w} + o_N(1),
    \end{equation}
    where $o_N(1)$ vanishes as $N\to\infty$, and where $\varphi : (-1,1) \to \R$ is defined as
    \begin{equation} \label{eq:varphi}
        \varphi(a) = \E_{X_1 \mid Y = 1}\left[ X_1 \operatorname{erf}\left( X_1 \operatorname{alg}^{-1}(a) / \sqrt{2} \right)
        \right]
    \end{equation}
    and $\operatorname{alg}^{-1}(x) = x/\sqrt{1-x^2}$, the inverse of the algebraic sigmoid function $\operatorname{alg}(x) = x/\sqrt{1+x^2}$.
\end{lemma}
Lemma~\labelcref{lem:gradient_flow} reduces the study of higher-order statistics to the marginal distributions, $X_1$, where, by translation invariance, all marginals have the same distribution, so we refer to $X_1$ without loss of generality.
While Lemma~\labelcref{lem:gradient_flow} technically only holds early in training and breaks down if $\mathbf{w}$ becomes localized due to violation of \labelcref{item:lindeberg-condition}, the gradient flow in \cref{eq:gradient_flow_early} holds sufficiently long to detect the emergence of localization in the weights $\mathbf{w}$.
In particular, numerically integrating \cref{eq:gradient_flow_early} yields localized weights $\mathbf{w}$ as $t \to \infty$.
Moreover, the location of the peak of final weights from \cref{eq:gradient_flow_early} corresponds closely to the actual peak of the weight, when we observe localization; see \cref{sec:peak-prediction} 
for empirical validation of this fact.
The primary difference observed is that the localized bump from \cref{eq:gradient_flow_early} is less peaked than when computed exactly; see \cref{fig:theory} for a comparison between experimentally observed localized receptive fields and theoretical predictions.

\subsection{Necessary and sufficient conditions for emergent localization}
\label{subsec:localization_conditions}

To establish an exact threshold at which localization emerges requires solving \cref{eq:gradient_flow_early}, which is not possible exactly for general nonlinear differential equations.\smash{\footnotemark}\footnotetext{
We discuss a partial differential equation limit that faces similar intractabilities in \cref{sec:pde-limit}.
}
Nevertheless, the form of \cref{eq:gradient_flow_early} reveals that localization is driven solely by the first term. 
Indeed, the second term depends only on the second-order statistics of the data, and so can be held fixed as $\mathbf{X}$ is varied from a distribution that induces localization to one that does not.
Secondly, one can see that the first term in \cref{eq:gradient_flow_early} does not change as $\mathbf{w}$ is scaled, in contrast to the second term.
As such, the second term in \cref{eq:gradient_flow_early} serves to constrain the \emph{scale} of $\mathbf{w}$, distinct from localization, while the first is primarily concerned with the \emph{shape} of $\mathbf{w}$, and thus localization.
This further motivates the first term, and thus $\varphi$, 
which we will refer to as the \emph{amplifier} and which itself depends on properties of the data distribution $p(\mathbf{X})$,
as a focus of study for understanding localization.

\renewcommand{\marginalheight}{25pt}
\newcommand{\varphiheight}{38pt}
\newcommand{\fieldheight}{48pt}
\begin{figure}[t]
\centering
    \hspace{-2em}
    \begin{tabular}{p{51pt}@{\hspace{25pt}}m{1pt}c@{\hspace{15pt}}m{1pt}c@{\hspace{15pt}}m{1pt}c}
      \raisebox{23pt}{
      \centering
      \begin{tabular}{c}
        \small$\texttt{Ising}$ \\
        \hspace{5pt}
        \includegraphics[height=\marginalheight]{figures/task/marginal/ising.pdf}
      \end{tabular}} &
      \raisebox{48pt}{\rotatebox{90}{\tiny $\varphi(a)$}} &
      \raisebox{8pt}{\includegraphics[height=\varphiheight]{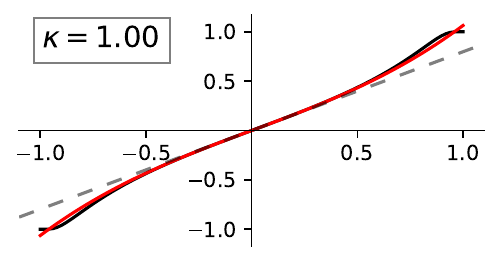}} & 
      \raisebox{44pt}{\rotatebox{90}{\tiny magnitude $w_i$}} &
      \includegraphics[height=\fieldheight]{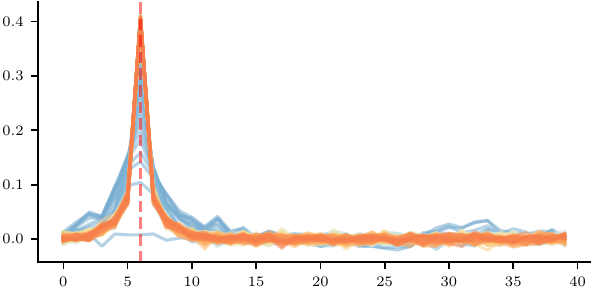} & 
      \raisebox{44pt}{\rotatebox{90}{\tiny magnitude $w_i$}} &
      \includegraphics[height=\fieldheight]{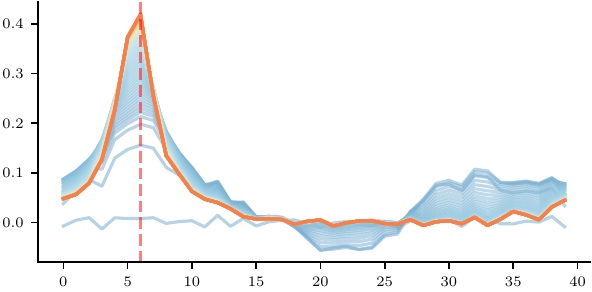} \\ 
      \noalign{\vskip -25pt}
      \raisebox{23pt}{
      \begin{tabular}{c}
        \small$\texttt{NLGP}(0.01)$ \\
        \hspace{4pt}
        \includegraphics[height=\marginalheight]{figures/task/marginal/gaussian.pdf}
      \end{tabular}} &
      \raisebox{48pt}{\rotatebox{90}{\tiny $\varphi(a)$}} &
      \raisebox{8pt}{\includegraphics[height=\varphiheight]{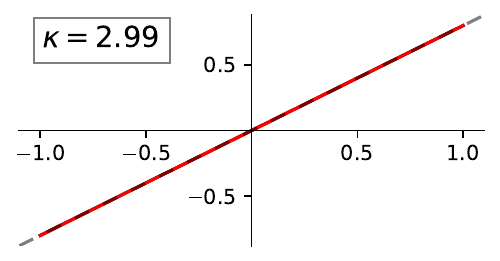}} &
      \raisebox{48pt}{\rotatebox{90}{\tiny magnitude $w_i$}} &
      \includegraphics[height=\fieldheight]{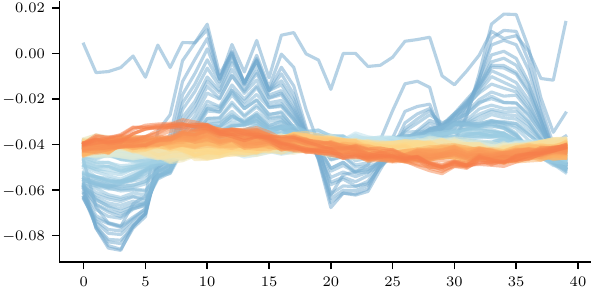} &
      \raisebox{48pt}{\rotatebox{90}{\tiny magnitude $w_i$}} &
      \includegraphics[height=\fieldheight]{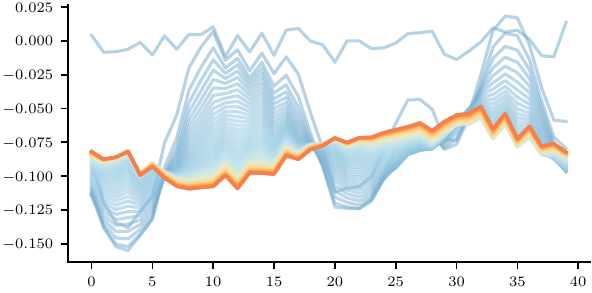} \\ 
      \noalign{\vskip -25pt}
      \raisebox{23pt}{
      \hspace{.4em}
      \begin{tabular}{c}
        \small $\texttt{Kur}(5)$ \\
        \includegraphics[height=\marginalheight]{figures/task/marginal/alg5.pdf}
      \end{tabular}} &
      \raisebox{48pt}{\rotatebox{90}{\tiny $\varphi(a)$}} &
      \raisebox{8pt}{
      \includegraphics[height=\varphiheight]{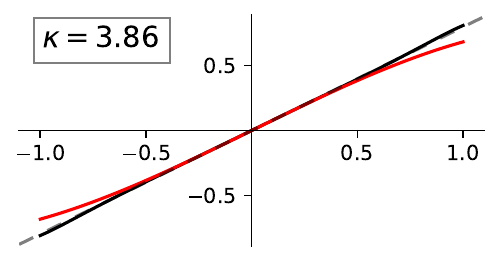} 
      } &
      \raisebox{44pt}{\rotatebox{90}{\tiny magnitude $w_i$}} &
      \includegraphics[height=\fieldheight]{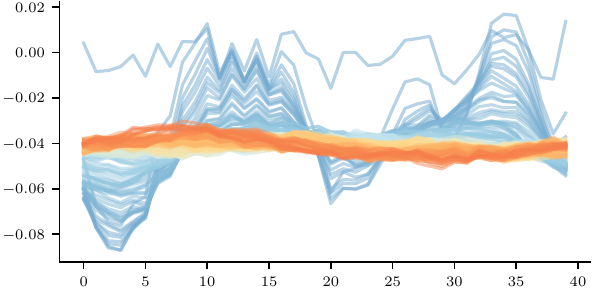} &
      \raisebox{44pt}{\rotatebox{90}{\tiny magnitude $w_i$}} &
      \includegraphics[height=\fieldheight]{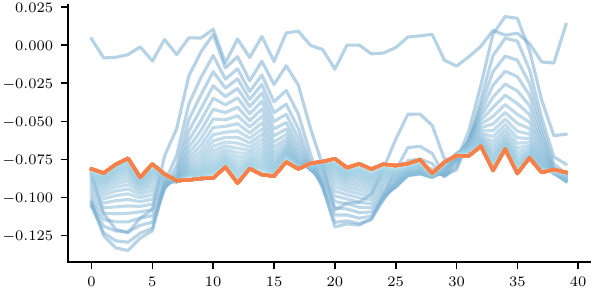} \\ 
      \noalign{\vskip -40pt}
      &&
      \hspace{0pt}\tiny input value $a$ & &
      \hspace{12pt}\tiny dimension $i$ of weight $\mathbf{w}$ & &
      \hspace{12pt}\tiny dimension $i$ of weight $\mathbf{w}$ \\
    \end{tabular}
  \caption{
    From left and for the same \texttt{Ising}, \texttt{NLGP}, and \texttt{Kur} data models as in \cref{fig:task}:
    the marginals $p(X_i)$,
    the amplifier $\varphi$ defined in \cref{lem:gradient_flow} and kurtosis $\kappa$,
    and the evolution of simulated receptive fields for the single-neuron model (\labelcref{item:single-neuron-model}) trained on its data, and 
    lastly the receptive field given by numerically integrating \cref{eq:gradient_flow_early} with $\varphi$ expanded to a third-order Taylor approximation for the same data;
    training or evolution time is indicated by line color (blue for early-time; red for late-time).
    \emph{See \cref{sec:theory-validation} for exposition.}
  }
  \label{fig:theory}
\end{figure}

We present an analysis of $\varphi$ in \cref{sec:varphi-analysis} that reveals the role of the marginal distribution of the data in driving localization.
For each marginal, $\varphi(a) \approx (\sqrt{2/\pi}) a$ for $a \approx 0$.
For larger $a$, $\varphi$ depends more strongly on the data distribution and can be super-linear (sub-linear), \ie greater (smaller) than $(\sqrt{2/\pi}) a$.
Super-linear $\varphi$ encourage entries in $\mathbf{w}$ that are large in some neighborhood to grow faster than those that are smaller, yielding localization.
Linear and sub-linear $\varphi$ are the opposite, encouraging oscillatory or flat weights by suppressing neighborhoods in $\mathbf{w}$.
However, super- and sub-linearity may not hold uniformly, as $\varphi$ can be both  over its domain (see \cref{fig:theory}, bottom row, black line).
As an approximation, we consider a third-order Taylor expansion (red lines in \cref{fig:theory}, second column), which reveals that for the canonical setting of $\sigma^2 = 1$, \emph{negative excess kurtosis of the marginals yields super-linearity}, while \emph{positive excess kurtosis yields sub-linearity};
see \cref{sec:varphi-analysis}.
This leads us to the following claim, which is validated by our simulations in \cref{sec:experiments}:
\begin{claim} \label{thm:localization}
    For sufficiently large $N$, if the data $\mathbf{X} \in \R^N$ satisfies conditions \labelcref{item:weak-dependence,item:translation-invariance,item:sign-symmetry} and has marginal distributions with sufficiently \emph{negative} excess kurtosis, then Model~\labelcref{item:single-neuron-model} will learn localized receptive fields.
    Conversely, if the excess kurtosis is sufficiently \emph{positive}, it will not.
\end{claim}

As a minimal positive example, the distribution with the most negative excess kurtosis is the symmetric Bernoulli, with a value of $-2$.
In our setting, this corresponds to a data vector $\mathbf{X}$ with support on the vertices of the hypercube, $\{ \pm 1 \}^N$.
As mentioned above, it can be seen from the law of total covariance combined with sign-symmetry that \labelcref{item:mean-assumption,item:covariance-assumption} hold exactly. %, \ie they are not assumptions.
Note that $\varphi$ is the same for all such distributions, which leads us to Claim~\labelcref{thm:localization} that \emph{any} distribution satisfying conditions \labelcref{item:weak-dependence,item:translation-invariance,item:sign-symmetry} whose marginals are maximally concentrated will induce a localized receptive field in~\labelcref{item:single-neuron-model}.
Importantly, this claim includes the limiting case of \textcite{ingrosso2022data}, $g\to\infty$ in $\texttt{NLGP}$.
It also includes the Ising model as another example, corroborating an observation for restricted Boltzmann machines \cite{harsh2020placecell} that Ising data induces localization in a learning model.
These claims are validated for the single-neuron model in \cref{fig:theory} and in \cref{sec:experiments} for the many-neuron model.

\subsection{Case study: Elliptical distributions fail to produce localization}
\label{sub:elliptical}

Above, we assume weak dependence (\labelcref{item:weak-dependence}) as it enables a focus on how the marginals control localization.
As a first investigation into departures from this regime, we consider data $\mathbf{X}$ sampled from an elliptical distribution, where weak dependence may not hold.
We specialize the definition of an elliptical distribution~\parencite{frahm2004generalized} to our setting of multiple class labels and sign-symmetry:
\begin{definition}
    \label{def:elliptical}
    Samples $\mathbf{X} \in \R^N$ satisfy an \emph{elliptical distribution} if we can write
    $\mathbf{X} \mid Y = y \overset{(d)}{=} R_y \Lambda_y \mathbf{U}_y$
    where $R_y$ is a nonnegative random variable, $\Lambda_y \in \R^{N \times D}$ is such that $\Lambda_y \Lambda_y^\top = \Sigma_y$, and $\mathbf{U}_y$ is independent of $R_y$ and uniformly distributed on the $D$-dimensional sphere.
\end{definition}

The class of elliptical distributions is broad, imposing only the constraint that the contours of the density be ellipses;
the multivariate Gaussian and Student-$t$ distributions are examples. % of elliptical distributions.
As such, they can vary greatly in measures of non-Gaussianity, including kurtosis, while maintaining enough structure for analytical convenience.
Proposition \labelcref{thm:elliptical} states that training on elliptical data \emph{prevents} localization in the single ReLU neuron model.
\begin{proposition} \label{thm:elliptical}
  Assume the data $\mathbf{X}$ are 
  sign-symmetric (\labelcref{item:sign-symmetry}),
  translation-invariant (\labelcref{item:translation-invariance}), 
  and follow an elliptical distribution such that the MSE on the task in \cref{sec:task} is always finite.
    If $\Sigma_0, \Sigma_1$ are such that the ratio of their $i$-th eigenvalues, $\lambda_i(\Sigma_0) / \lambda_i(\Sigma_1)$, assumes a particular value for at most two distinct $i$, then the steady states of the weight of \labelcref{item:single-neuron-model} are sinusoids, \ie not localized.
\end{proposition}

The condition on the number $i$ such that the ratio of the $i$-th eigenvalues are the same constrains the number of Fourier components that can be non-zero in the steady state of $\mathbf{w}$.
While opaque, this requirement seems to always hold in practice, as even slight increases in length-scale correlation can dramatically change the spectrum of $\Sigma_y$.

The proposition is surprising because it reveals that the kurtosis of the preactivation is not an appropriate metric for explaining localization.
Consider the example of the $N$-dimensional Student-$t$ distribution with $\nu$ degrees of freedom, $t_N(\nu)$.
If $\mathbf{X} \sim t_N(\nu)$, then $\langle w, \mathbf{X} \rangle \sim t_1(\nu)$.
Note the kurtosis of $t_1(\nu)$ is non-zero, and can be very large or even infinite for small $\nu$.
This prediction is validated in \cref{sec:elliptical-experiments}.
The condition also reveals that not all symmetries in the data (here, elliptical symmetry) induce structure in the trained model weights, if localization is to be seen as a sparsity more structured than oscillatory weights~\parencite[\cf][]{godfrey2023symmetries}; indeed, translational symmetry (\labelcref{item:translation-invariance}) is more relevant for localization than elliptical symmetry.

\section{Experimental results}
\label{sec:experiments}

We describe experiments to validate the generalizability of the analytical results from \cref{sec:theory}.
We run all experiments on a single CPU machine locally or on a compute cluster. 
Since all datasets are procedurally generated, training depends on both the model architecture and the complexity of sampling the data, 
but is between 10 and 60 minutes for any single simulation run.

\subsection{Validating Claim~\labelcref{thm:localization} with positive and negative predictions}
\label{sec:theory-validation}
\renewcommand{\fieldheight}{48pt}
\begin{figure}[t]
  \centering
  \begin{tabular}{m{1pt}c@{\hspace{20pt}}m{1pt}c@{\hspace{20pt}}m{1pt}c}
    &\hspace{12pt}\small $t_{40}(\nu=3)$ && \hspace{12pt}\small\texttt{elliptical shell} && \hspace{12pt}\small\texttt{elliptical extreme} \\
    \raisebox{48pt}{\rotatebox{90}{\tiny magnitude $w_i$}} &
    \includegraphics[height=\fieldheight]{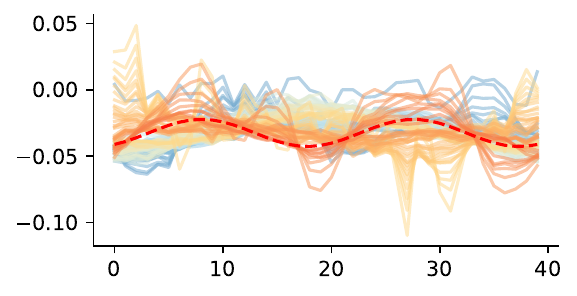} &
    \raisebox{48pt}{\rotatebox{90}{\tiny magnitude $w_i$}} &
    \includegraphics[height=\fieldheight]{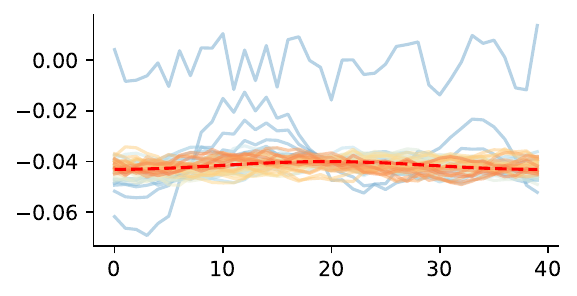} & 
    \raisebox{48pt}{\rotatebox{90}{\tiny magnitude $w_i$}} &
    \includegraphics[height=\fieldheight]{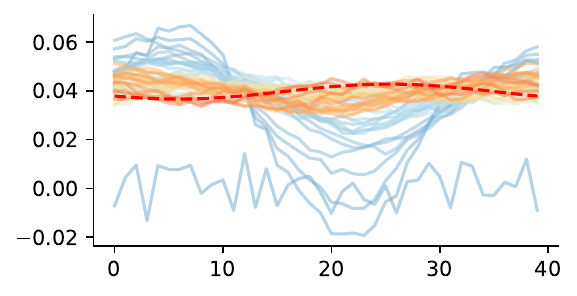} \\
    \noalign{\vskip -40pt}
    &
    \hspace{12pt}\tiny dimension $i$ of weight $\mathbf{w}$ & &
    \hspace{12pt}\tiny dimension $i$ of weight $\mathbf{w}$ &&
    \hspace{12pt}\tiny dimension $i$ of weight $\mathbf{w}$ 
  \end{tabular}
  \caption{
    Evolution of receptive fields learned by the single-neuron model (\labelcref{item:single-neuron-model}), along with sinusoids fit to final states (red dashes) when trained on data from three elliptical distributions: $t_{40}(\nu=3)$ (\textbf{left}), the surface of an ellipse (\textbf{middle}), and a custom elliptical distribution that places its mass near the outside of an ellipse (\textbf{right}).
    In all cases, the learned receptive field is oscillatory (a sinusoid), as predicted by Proposition \ref{thm:elliptical}.
    The $\ell_2$ distances between the fitted oscillatory weights and empirical RFs, as a ratio of the $\ell_2$ norm of the empirical RFs, are (left) 9.77\%, (center) 3.75\%, and (right) 4.14\%.
    \emph{See \cref{sec:elliptical-experiments} for exposition.}
    }
    \label{fig:elliptical}
\end{figure}

\begin{wrapfigure}{l}{0.45\textwidth}
    \centering
    \includegraphics[width=170pt]{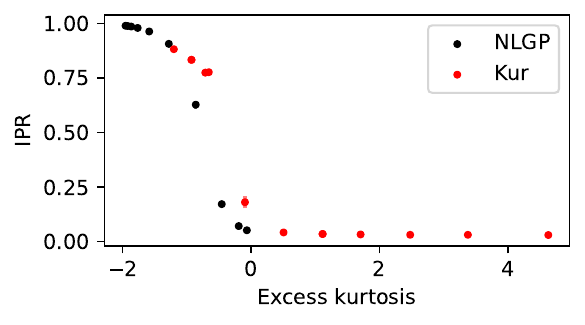}
    \caption{
    IPR \vs excess kurtosis for $\texttt{NLGP}$ and $\texttt{Kur}$ data models, with mean and std.~dev.~across 30 re-initializations for the single-neuron model (\labelcref{item:single-neuron-model});
    error bars are small and may not be visible.
    }
    \label{fig:replications}
    \vspace{-14pt}
\end{wrapfigure}

In \cref{fig:replications}, we validate Claim~\labelcref{thm:localization} first via the single-neuron model (\labelcref{item:single-neuron-model}) with 30 initial conditions trained across a range of excess kurtoses for the $\texttt{NLGP}(g)$ and $\texttt{Kur}(k)$ data models.
We use the inverse participation ratio (IPR), defined in \cref{app:IPR}.
This measure, also used by \textcite{ingrosso2022data}, is large when proportionally few weight dimensions ``participate'' (have large magnitude), and small when weight dimension magnitudes are more uniform.
We see that when $g$ and $k$ assume values that yield a negative excess kurtosis, IPR is close to its maximum of $1.0$, suggesting the weights are localized; if the excess kurtosis is positive, IPR is nearly zero, suggesting the weights are \emph{not} localized.
The IPR is extremely consistent across random initializations, suggesting that localization is determined by data statistics and not initialization.
The trend in IPR \vs excess kurtosis is very similar between the $\texttt{NLGP}(g)$ and $\texttt{Kur}(k)$ data models, demonstrating that excess kurtosis is a primary driver of localization and localization is largely independent from other properties of the data distribution.

\Cref{fig:theory} further validates Claim~\labelcref{thm:localization} with specific examples.
We maintain constant initial conditions for our model and train on the \texttt{Ising}, $\texttt{NLGP}(g=0.01)$, and $\texttt{Kur}(k=5)$ data models.
The marginals of the \texttt{Ising} model have an excess kurtosis of $-2$, the smallest possible value for any distribution.
As a result, we see that the amplifier $\varphi$ for \texttt{Ising} (top left) is super-linear (the dark line exceeds the dashed light line for larger $a$), which drives localization via its role in \cref{eq:gradient_flow_early}.
Integrating \cref{eq:gradient_flow_early} with $\varphi$ expanded via a third-order Taylor approximation (red line) yields a similar localized receptive field to that from simulation (two right panels), validating this approximation.

For the remaining distributions (middle and bottom rows) that elicit oscillatory (sinusoidal) weights, 
Claim~\labelcref{thm:localization} is validated due to their positive excess kurtosis.
The dynamical steady state (far right) assumes a more negative value than in the simulation (to the left), 
a difference that is the result of deviations of our \emph{early-time} gradient flow in 
\cref{eq:gradient_flow_early}, but these deviations remain mild enough nevertheless to recover the qualitative structure of the learned receptive field.

\begin{figure}[t]
  \scalebox{0.92}{
    \begin{tabular}{m{1pt}c@{\hspace{10pt}}c@{\hspace{10pt}}c}
      &\hspace{7pt}\small model: \labelcref{item:many-neuron-model}; data: $\texttt{Kur}(10)$) &
      \hspace{7pt}\small model: \labelcref{item:many-neuron-model}; data: $\texttt{Kur}(4)$ &
      \hspace{7pt}\small model: ICA; data: $\texttt{Kur}(3)$ \\
      \raisebox{62pt}{\rotatebox{90}{\tiny magnitude $w_i$}} &
      \includegraphics[width=0.33\linewidth]{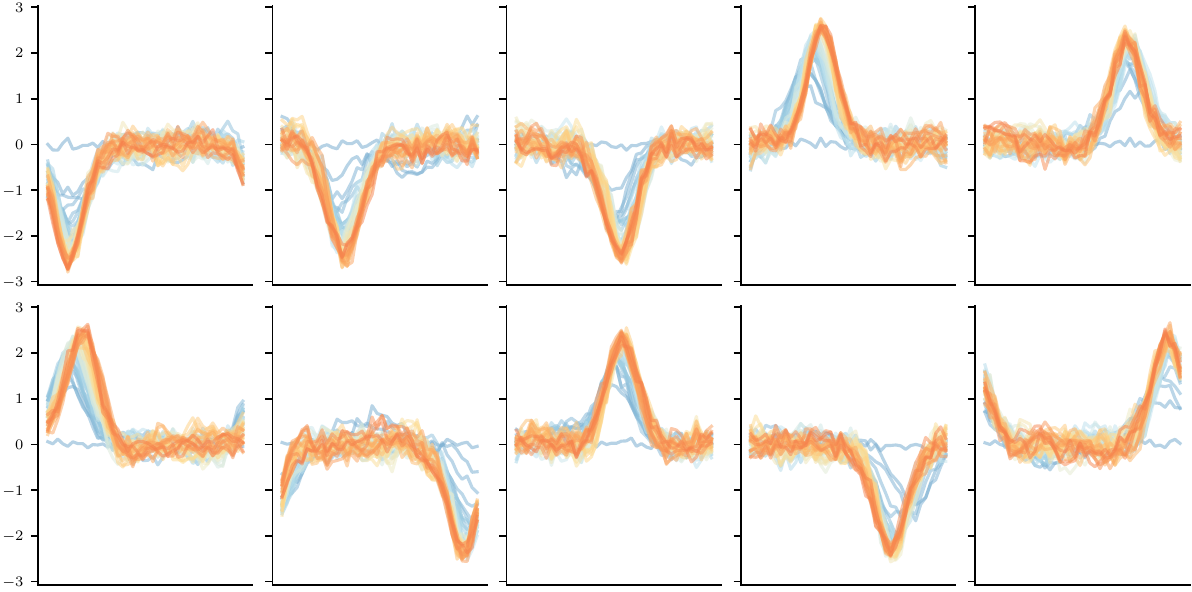} &
      \includegraphics[width=0.33\linewidth]{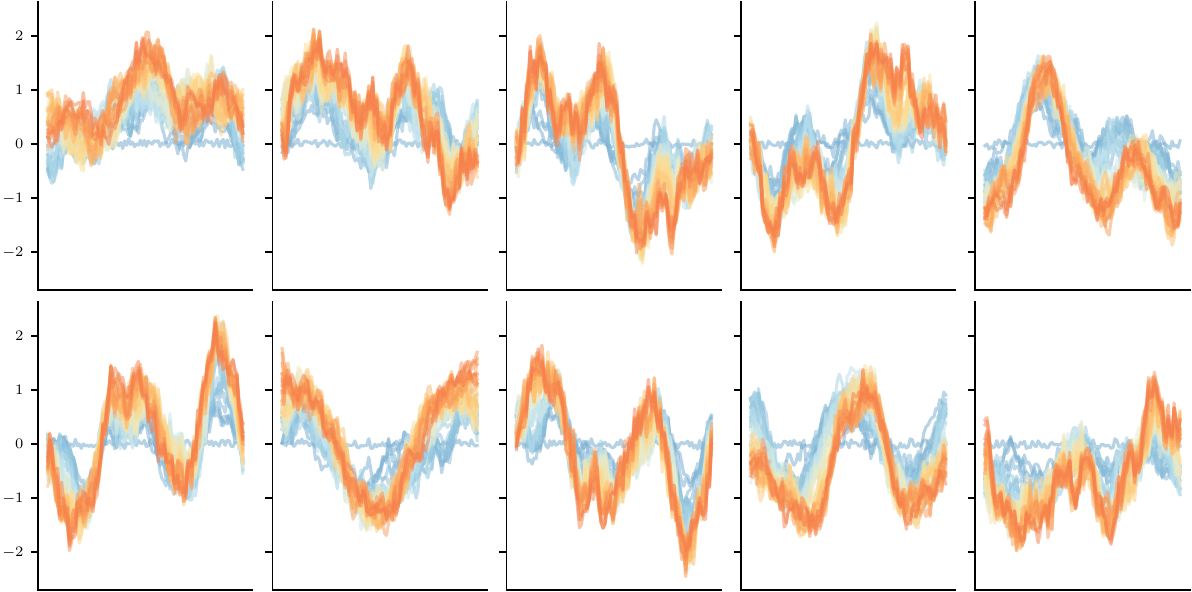} &
      \includegraphics[width=0.33\linewidth]
      {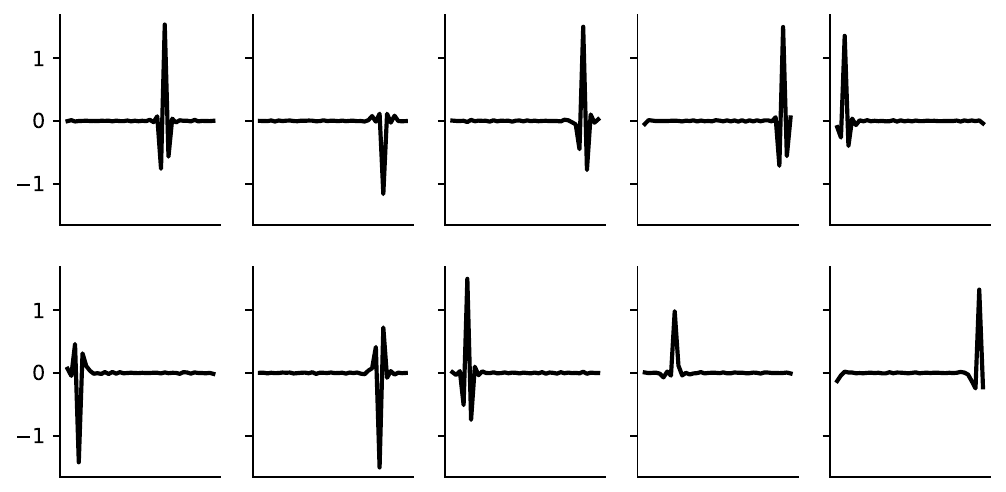} \\
    \noalign{\vskip -48pt}
      &\hspace{7pt}\tiny dimension $i$ of weight $\mathbf{w}$ &
      \hspace{7pt}\tiny dimension $i$ of weight $\mathbf{w}$ &
    \hspace{7pt}\tiny dimension $i$ of weight $\mathbf{w}$ 
    \end{tabular}
  }
  \caption{
    (\textbf{Left}, \textbf{Center}) Receptive fields learned 
    by many-neuron (\labelcref{item:many-neuron-model}) 
    soft committee machines (second-layer weights fixed at $\frac{1}{K}$) 
    trained on the $\texttt{Kur}(10)$ and $\texttt{Kur}(4)$ datasets, respectively.
    The models had $N=40$ input units, $K=10$ hidden units, and an initialization variance of $0.1$.
    (\textbf{Right}) A random subset of 10 components from the 40 learned by 
    the FastICA algorithm from scikit-learn~\parencite{hyvarinen1999fast,scikit-learn} on the 
    $\texttt{Kur}(3)$ dataset with 
    length-scale correlation values of $\xi_0 = 1$ and $\xi_1 = 3$. 
    \emph{See \cref{sec:extensions} for exposition.}
  }
  \label{fig:extensions}
\end{figure}

\subsection{Validating \cref{eq:gradient_flow_early} with localization position prediction}
\label{sec:peak-prediction}
The simulated and integrated receptive fields in \cref{fig:theory} demonstrate that our analytical model is able to meaningfully reproduce localization in receptive fields from neural network training.
For the Ising model, we see that the integration even has a peak in the exact same position as the simulation (at index $i=6$), suggesting precision in our approximation.
Indeed, we simulated the condition in \cref{fig:theory} for the Ising model under 28 different initial conditions (weight initializations), and found that in 26 of them (93\%), the peaks of the integrated and simulated receptive fields matched exactly.
In the two cases where the peaks differed, they did so substantially (see \cref{fig:time} for an example). 

\subsection{Validating Proposition \ref{thm:elliptical}: Elliptical distributions fail to localize}
\label{sec:elliptical-experiments}
Proposition \ref{thm:elliptical} claims that the single-neuron model (\labelcref{item:single-neuron-model})
trained on elliptical data will yield sinusoidal receptive fields, 
subject to a condition on the spectra of $\Sigma_0$ and $\Sigma_1$.
We verify this claim in \cref{fig:elliptical} with three distinct elliptical distributions.
The first, $t_{40}(\nu=3)$, gives preactivations $\langle \mathbf{w}, \mathbf{X} \rangle$ that have \emph{infinite} kurtosis, yet our theory predicts the final receptive field will be sinusoidal.
This is confirmed in \cref{fig:elliptical}, where the learned receptive field is indeed a sinusoid with period 1 and intercept at zero.

We also consider data sampled from the surface of an ellipse, which is done by fixing $R_y \equiv 1$ in \cref{def:elliptical}.
Here, we observe that the learned receptive field is a near-constant function at $-0.04$ (note that $\cos(2\pi \cdot 0 \cdot x) \equiv 1$ is a sinusoid, allowing nonzero intercepts and constant functions).
Finally, we consider an unconventional elliptical distribution where the density of $R$ is given by
$p_R(r) = (4e^{2r+4}) / (e^{2r}+e^{4})^{2} \cdot \mathbbm{1}(r \geq 2)$.
This particular density places most of its mass near $r = 2$ before rapidly falling off, imposing a minimum norm on $\mathbf{X}$ and pushing support near the surface of an ellipse.
This distribution, too, yields an oscillatory steady state, as shown in \cref{fig:elliptical} (right). 
We confirm our visual observations by fitting sinusoids to the final receptive fields and see the relative errors are quite low.

\subsection{Extensions to many-neuron model and ICA}
\label{sec:extensions}

All of our analysis thus far has concerned single-neuron models with ReLU activation and without hidden-to-output or bias terms, assumptions which were made to make our analysis tractable.
Here, we depart from that regime by considering the SCM and the full two-layer network (Model~\labelcref{item:many-neuron-model}).
In \cref{fig:extensions} (left) and (center), we train a SCM with 10 hidden units and sigmoid activation on the $\texttt{Kur}(10)$ and $\texttt{Kur}(4)$ datasets, which have excess kurtoses of $-0.93$ and $3.28$, respectively.
So, based on our single-neuron analysis, we \emph{do} and \emph{do not} expect to see localization for these distributions.
Indeed, this is precisely what we observe in \cref{fig:extensions}, where the receptive fields are sharply localized for the former distribution, while they look like low-frequency oscillations for the latter.

\begin{wrapfigure}{l}{0.45\textwidth}
  \scalebox{0.9}{
    \hspace{-8pt}
    \begin{tabular}{m{1pt}c}
      \multirow{2}*{
        \rotatebox{90}{\tiny magnitude $w_i$}
      }
      \hspace{2pt}
      &
      \includegraphics[width=\linewidth]{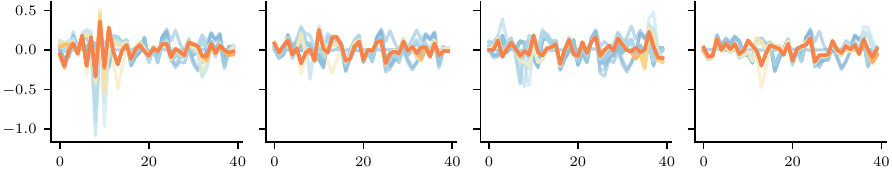} \\ 
      &
      \includegraphics[width=\linewidth]{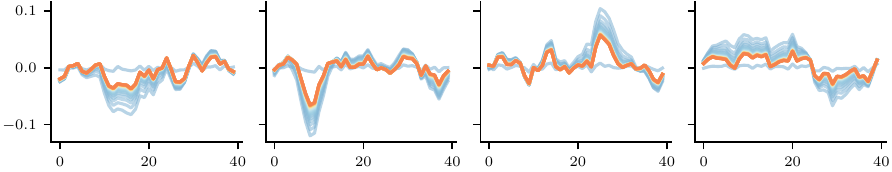}  \\
      \noalign{\vskip -4pt}
      &\hspace{18pt}\tiny dimension $i$ of weight $\mathbf{w}$
    \end{tabular}
  }
  \caption{
    Receptive fields learned by the many-neuron model (\labelcref{item:many-neuron-model})
    with learnable second-layer weights, $N=40$, $K=10$.
    (\textbf{Top}) A random subset of 4 receptive fields from a model with sigmoid activation, 
    trained on $\texttt{Kur(4)}$ (positive excess kurtosis of $3.28$).
    As predicted by Claim~\labelcref{thm:localization},
    the receptive fields are \emph{not} localized, and appear as high-frequency oscillations.
    (\textbf{Bottom}) A random subset of 4 receptive fields from a model with ReLU activation,
    trained on $\texttt{Kur(30)}$ (negative excess kurtosis of $-1.17$).
    Receptive fields are localized (\textbf{left three}) or exhibit low-frequency oscillations (\textbf{right}).
    \emph{See \cref{sec:extensions} for exposition.} 
  }
  \label{fig:multi-neuron}
\end{wrapfigure}

In \cref{fig:multi-neuron}, we train many-neuron models with $N=40$ input units and $K=10$ hidden units, where all weights are learnable.
In general, adding flexibility in the second layer leads to more varied structure in the first layer.
We train on $\texttt{Kur}(4)$ (top), which has an excess kurtosis of $3.28$, and $\texttt{Kur}(30)$ (bottom), which has an excess kurtosis of $-1.17$.
The receptive fields from the former are not localized, as in the single-neuron model; however, they appear more like high-frequency oscillations than low-frequency sinusoids.
For $\texttt{Kur}(30)$, where we expect localization, we see that the first three receptive fields exhibit localization, but less so than for a single neuron.
Importantly, not all receptive fields are localized,
a result of a variable second-layer weight effectively changing the variance $\sigma^2$ in the third-derivative term in Lemma~\labelcref{lem:varphi}.

We further compare these predictions against ICA, another framework that has been used to model receptive fields in visual cortex.
We train on the $\texttt{Kur}(3)$ dataset, which has marginals with excess kurtosis $7.66$, fitting 10 components using the FastICA implementation from scikit-learn \parencite{hyvarinen2000independent,scikit-learn}.
We observe in \cref{fig:extensions} (right) that we learn localized receptive fields; this contrasts our neural network models, which require negative excess kurtosis.
This stems from ICA's objective to maximize non-Gaussianity, regardless of how specifically it is done.
The sign of the excess kurtosis is irrelevant, so long as it is nonzero.
This deviation between our analytical model and ICA is an interesting avenue for future study, perhaps by validation with natural images.

\section{Conclusions}
\label{sec:conclusions}

We derive effective learning dynamics for the minimal example of emergent localization in a neural receptive field given by \textcite{ingrosso2022data}. 
The analytical approach we take relies on the assumption that the \emph{conditional} preactivation is Gaussian, a refinement of previous work that assumes Gaussianity of the unconditioned preactivation as asserted by the \emph{Gaussian equivalence property} targeted by \textcite{ingrosso2022data}.
This approach may prove extensible beyond our specialized setting and may enable further analysis of how statistics of an input task drive emergent structure in neural network learning.

Emergence as an alternative mechanism to top-down constraints like sparsity
is in line with recent work that reformulates data-distributional properties as a driver for complex behavior~\parencite{chan2022data}. 
Via these analytical effective dynamics, we observe that specific data properties---the covariance structure 
and the marginals---shape localization in neural receptive fields.
Though we cannot capture dynamical interactions between neurons that may shape receptive fields in other settings with the single-neuron analytical model, our empirical validations with many neurons suggest that these interactions do not, in fact, play a significant role in shaping localization~\cite[\cf][]{harsh2020placecell}.

The data model we consider is a simplified abstraction of the task faced by early sensory systems, and, as a consequence, we do not yet capture certain features of receptive fields that are observed in early sensory systems.
In particular, we do not observe orientation nor phase selectivity, features of simple-cell receptive fields in early sensory cortices and in artificial neural networks that can be seen in a subset of receptive fields in \cref{fig:sim-real-gabors} (left and center, respectively).
To capture orientation selectivity, it may be fruitful to follow the approach of \textcite{karklin2011efficient}, who tie orientation selectivity in a population-based efficient-coding framework to the presence of noise.
Furthermore, on-center-off-surround-filtering input data, including the idealized data, gives receptive fields with subfields in our simulations, but is difficult to analyze.
Lastly, we do not yet look at the distribution of receptive field shapes and do not validate against other models of receptive field learning beyond a brief comparison with ICA~\parencite[\cf][]{saxe2011unsupervised}, but these are exciting avenues for future work.

\clearpage
\section*{Acknowledgements}
This work was supported by a Schmidt Science Polymath Award to A.S., and the Sainsbury Wellcome Centre Core Grant from Wellcome (219627/Z/19/Z) and the Gatsby Charitable Foundation (GAT3850). A.S. is a CIFAR Azrieli Global Scholar in the Learning in Machines \& Brains program.

\printbibliography

\clearpage
\appendix
\section{Definitions and Notation}

\subsection{Notation}
We use $[n]$ to refer to the set $\{ i \in \N : 1 \leq i \leq n \}$.

\subsection{Algebraic sigmoid}
\label{sec:algebraic-sigmoid}
For $k > 0$, the generalized algebraic sigmoid function is defined as
\begin{equation}
    \operatorname{alg}_k(x) \triangleq \frac{1}{2} \left( 1  + \frac{x}{(1+|x|^k)^{1/k}} \right)~.
\end{equation}
Following the main text, we drop the subscript when $k = 2$.

\subsection{Inverse participation ratio (IPR)}
\label{app:IPR}
The IPR is defined as:
$$ \operatorname{IPR}(\mathbf{w}) \triangleq \left(\sum_{i=1}^D w_i^4\right)/\left(\sum_{i=1}^D w_i^2\right)^2, $$
where $w_i$ is the magnitude of dimension $i$ of weight $\mathbf{w}$.

\section{Extensions beyond the scope of the main text}

\subsection{Analytical properties of the amplifier $\varphi$}
\label{sec:varphi-analysis}

We present several properties of the amplifying function $\varphi$
defined in Lemma~\labelcref{lem:varphi}.

\begin{lemma} \label{lem:varphi}
    The localization amplifier $\varphi$ in \cref{eq:varphi} satisfies $\varphi(-a) = -\varphi(a)$, for all $a \in (-1,1)$.
    Moreover, its derivatives satisfy, where $\sigma^2$ and $\kappa$ denote the variance and kurtosis of $X_1$, respectively:
    \vspace{-10pt}
    \begin{align*}
      \varphi'(0) &= \sqrt{\frac{2}{\pi}} \sigma^2~, &&\text { and }&&
      \varphi'''(0) = -\sqrt{\frac{2}{\pi}} (\kappa^4 \sigma^4 - 3 \sigma^2)~.
    \end{align*}
    \end{lemma}
\vspace{-6pt}

To gain some understanding of how the marginal distributions of $\mathbf{X}$ impact localization, we use the derivatives in Lemma~\labelcref{lem:varphi} 
to construct a third-order Taylor approximation of $\varphi$ about 0.
The derivatives in Lemma~\labelcref{lem:varphi} reveal that every distribution for $X_1$ with constant variance will look like the same linear function near 0.
$\varphi$ only looks nonlinear once we move sufficiently far away from zero when the third-order term becomes relevant.
For the case of $\sigma^2 = 1$ (where the variance of $X_1$ is equal to the value of the larger target $y=1$) the third order term suggests that $\varphi$ is super-linear when $\kappa < 3$, \ie the excess kurtosis is positive, and sub-linear otherwise.

A super-linear $\varphi$ will encourage entries where $\Sigma_1 \mathbf{w}$ is large to grow at a faster rate than other entries, which are all subject to the same \emph{linear} norm constraint through the second term in \cref{eq:gradient_flow_early}.
The covariance $\Sigma_1$, as a circulant matrix, acts as the convolution operator between some vector and $\sigma_1^1$ (the first row in $\Sigma_1$).
Since $y=1$ corresponds to the class with a larger length-scale correlation, $\sigma_1^1$ will decay relatively slowly and act like a weight local average.
Thus, $\Sigma_1 \mathbf{w}$ is the weighted local average for each entry in $\mathbf{w}$.
So, entries where $\mathbf{w}$ is large within some neighborhood will be encouraged to grow faster than those which are smaller, an effect that compounds as \cref{eq:gradient_flow_early} is integrated.
Thus, super-linearity encourages localization.

As we will see in \cref{thm:elliptical} for the setting of elliptical data, if $\varphi$ is linear, $\mathbf{w}$ learns to be sinusoidal, and thus not localized.
In the case of sub-linearity, we expect suppression of larger values, rather than promotion, as in the super-linear setting.
Thus, to a first approximation, the sign of the excess kurtosis, $\kappa - 3$ (for $\sigma^2 = 1$), indicates whether $\mathbf{w}$ localizes.

However, simply studying $\varphi'''(0)$ is not sufficient to fully characterize how the marginals impact localization.
A function can be sub-linear for small $a$ and super-linear for larger $a$, making it unclear whether it will yield localization.
For marginal distributions where $\kappa \approx 3$ that do not exhibit strict super- or sub-linearity, this condition is no longer precise enough to determine whether we see localization.

\subsection{PDE limit of \cref{eq:gradient_flow_early}}
\label{sec:pde-limit}

By taking $N$ to be large and treating $w$ as a continuous function with respect to position, \ie $w \equiv w(x, t)$, one can treat \cref{eq:gradient_flow_early} as a partial differential equation (PDE). 
Finding its steady state then amounts to solving
\begin{equation}
    \varphi\left( \frac{\sigma^1 \star w}{\sqrt{ \langle \sigma^1 \star w, w \rangle}} \right) - \frac{1}{2} (\sigma^0 + \sigma^1) \star w \equiv 0~,
\end{equation}
where $w : [0,1] \to \R$ is periodic and $\sigma^y$ is the convolution corresponding to the limiting case of the matrix $\Sigma_y$.
This equation does not appear to have an explicit solution for non-identity $\Sigma_1$,
and thus, it may not be possible to solve the steady states of \cref{eq:gradient_flow_early} exactly in this PDE limit or for finite $N$.

\subsection{Assumptions~\labelcref{item:mean-assumption,item:covariance-assumption}
\vs Gaussian equivalence}

Assumptions~\labelcref{item:mean-assumption,item:covariance-assumption} are equivalent to approximating $\langle \mathbf{w}, \mathbf{X} \rangle \mid X_i$ as Gaussian early in training.
Similar ideas have been used to derive gradient flow dynamics for neural networks, including in developing the Gaussian equivalence property of \cite{goldt2020modelling,gerace2020generalisation,goldt2022gaussian}.
However, these works model the unconditional preactivation $\langle \mathbf{w}, \mathbf{X} \rangle$ as a Gaussian, rather than first conditioning on $X_i$.
How this arises is that these previous works assert the Gaussian approximation \emph{prior} to differentiating the loss function for the gradient flow dynamics.
However, an approximation at that stage neglects the presence of a multiplicative factor $\mathbf{X}$ that appears as a result of the chain rule applied to $\langle \mathbf{w}, \mathbf{X} \rangle$.
Abstractly, this approach assumes that $\LL_\text{exact} \to \LL_\text{Gauss}$ implies $\nabla_\mathbf{w} \LL_\text{exact} \to \nabla_\mathbf{w} \LL_\text{Gauss}$, but, in general, this does not follow, and here in particular this assumption does not capture the interplay of learning and higher-order input statistics. %\nb{Can I say more?}
This contributes to the failure of Gaussian equivalence in \cite{ingrosso2022data}.
In contrast, we can account for the additional $\mathbf{X}$ term in the derivation of 
Lemma~\labelcref{lem:gradient_flow} by assuming that $\langle \mathbf{w}, \mathbf{X} \rangle \mid X_i$ rather than $\langle \mathbf{w}, \mathbf{X} \rangle$ is Gaussian.
This conditioning approach, along with the translation invariance of the data (Property~\labelcref{item:translation-invariance}), also motivates considering the marginal distributions $X_i$ as the object of study to obtain gradient flows for neural networks trained on non-Gaussian inputs.

\section{Proofs of theoretical results}
\label{app:proofs}

\subsection{Gradient flow for mean-squared error (MSE) loss}
The loss is given by:
\begin{align}
    \LL 
    &= \E_{\mathbf{X},Y}\left[ (Y - \operatorname{ReLU}(\langle \mathbf{w}, \mathbf{X} \rangle) )^2 \right] \notag \\
    &= \E_{\mathbf{X},Y}\left[ Y^2 \right] - 2 \underbrace{ \E_{\mathbf{X},Y}\left[ Y \operatorname{ReLU}(\langle \mathbf{w}, \mathbf{X} \rangle) \right] }_{\triangleq (I)} + \underbrace{ \E_{\mathbf{X},Y}\left[ \operatorname{ReLU}^2(\langle \mathbf{w}, \mathbf{X} \rangle) \right] }_{\triangleq (II)}. \label{eq:loss_2relu_neuron}
\end{align}
The assumption of sign symmetry (\labelcref{item:sign-symmetry}) gives that 
$\langle \mathbf{w}, \mathbf{X} \rangle$ is also sign-symmetric.
First, this implies that $\PR( \langle \mathbf{w}, \mathbf{X} \rangle > 0 ) = \frac{1}{2}$, so:
\begin{align*}
    (II)
    &= \frac{1}{2} \E_{\mathbf{X},Y \mid \langle \mathbf{w}, \mathbf{X} \rangle \geq 0}\left[ \operatorname{ReLU}^2(\langle \mathbf{w}, \mathbf{X} \rangle) \right]
    = \frac{1}{2} \E_{\mathbf{X},Y \mid \langle \mathbf{w}, \mathbf{X} \rangle \geq 0}\left[ (\langle \mathbf{w}, \mathbf{X} \rangle)^2 \right].
\end{align*}
Second, sign-symmetry of $\langle \mathbf{w}, \mathbf{X} \rangle$ 
implies that we can drop the conditioning on $\langle \mathbf{w}, \mathbf{X} \rangle \geq 0$, since $\langle \mathbf{w}, \mathbf{X} \rangle \overset{d}{=} -\langle \mathbf{w}, \mathbf{X} \rangle$.
Thus,
\begin{align*}
    (II)
    &= \frac{1}{2} \E_{\mathbf{X},Y}\left[ (\langle \mathbf{w}, \mathbf{X} \rangle)^2 \right] \\
    &= \frac{1}{2} \mathbf{w}^\top \E_{\mathbf{X},Y} \left[ \mathbf{X} \mathbf{X}^\top \right] \mathbf{w} \\
    &= \frac{1}{2} \mathbf{w}^\top \left( \frac{1}{K} \sum_{y=0}^{K-1} \Sigma_y \right) \mathbf{w}  \\
    &= \frac{1}{2K} \sum_{y=0}^{K-1} \mathbf{w}^\top \Sigma_y \mathbf{w}~,
\end{align*}
where $K$ is the number of values (classes) of discrete $y$.
Finally, we differentiate $\LL$ with respect to $\mathbf{w}$:
\begin{align*}
  \nabla_\mathbf{w} \LL &= 2 \E_{\mathbf{X},Y}\left[ Y \mathbbm{1}( \langle \mathbf{w}, \mathbf{X} \rangle \geq 0) \mathbf{X} \right] + \frac{1}{K} \sum_{y=0}^{K-1} \Sigma_y \mathbf{w}~.
\end{align*}
The gradient flow~\cite{elkabetz2024continuous} is given by $\frac{\mathrm{d}\mathbf{w}}{\mathrm{d}t} = -\tau \nabla_\mathbf{w} \LL$, where $\tau$ is the learning rate.
Thus,
\begin{equation}
  \frac{1}{\tau} \frac{\mathrm{d}\mathbf{w}}{\mathrm{d}t}
    = 2 \E_{\mathbf{X},Y}\left[ Y \mathbbm{1}( \langle \mathbf{w}, \mathbf{X} \rangle \geq 0) \mathbf{X} \right] - \frac{1}{K} \sum_{y=1}^{K} \Sigma_y \mathbf{w}~. \label{eq:gradient_flow_two}
\end{equation}

\subsection{Proof of lemma \ref{lem:gradient_flow}} \label{subsec:pf_of_gradient_flow}
Setting $K = 2$ in equation \eqref{eq:gradient_flow_two}, we have
\begin{align*}
  \frac{1}{\tau} \frac{\mathrm{d}\mathbf{w}}{\mathrm{d}t} 
  &= \E_{\mathbf{X} \mid Y=1}[ \mathbbm{1}( \langle \mathbf{w}, \mathbf{X} \rangle \geq 0 ) \mathbf{X} ] - \frac{1}{2} \left( \Sigma_0 + \Sigma_1 \right) \mathbf{w}~.
\end{align*}
We wish to express the first term explicitly.
Note that the first term is a vector in $\R^N$.
We consider each of its entries separately by
using the law of total expectation to write the $i$-th entry as:
\begin{align*}
    \E_{\mathbf{X} \mid Y=1}[ \mathbbm{1}( \langle \mathbf{w}, \mathbf{X} \rangle \geq 0 ) X_i ] 
    &= \E_{ X_i \mid Y=1 } \left[ X_i \PR_{\mathbf{X} \mid X_i=x_i, Y = 1} \left[  \langle \mathbf{w}, \mathbf{X} \rangle \geq 0 \right] \right]~.
\end{align*}

By Assumption~\labelcref{item:lindeberg-condition}, $\{ w_i X_i \mid 1 \leq i \leq N \}$ satisfies Lindeberg's condition as $N\to\infty$.
This is also known as a \emph{uniform integrability} requirement.
Before formally stating it, let us introduce two variables: $S_N \triangleq \sum_{j=1}^{N} w_j (X_j - \mu_{j\mid x_i})$, the partial sums, and their variance, $\sigma_N^2 \triangleq \E[ S_N^2 ]$, where $\mu_{j \mid x_i} \triangleq \E[X_j \mid X_i = x_i]$ is the conditional mean of the $j$-entry given that $i$-th entry has value $x_i$.
Then, Lindeberg's condition is formally stated as
\begin{equation*}
  \sup_N \E_{S_N \mid X_i = x_i} \left[ \left| \tfrac{S_N^2}{\sigma_N^2} \right| \mathbbm{1} \left( \left| \tfrac{S_N^2}{\sigma_N^2} \right| > x \right) \right] \to 0 \quad \text{as} \quad x\to\infty~.
\end{equation*}
This condition effectively states that no term in the partial sum $w_i X_i$ will dominate.
Under this condition, along with weak dependence (Property~\labelcref{item:weak-dependence}), we conclude from \cite[Theorems 1.19, 10.2]{bradley2007introduction} that $S_N / \sigma_N \overset{d}{\to} \NN(0,1)$.
Note that
\begin{equation*}
    S_N = \langle \mathbf{w}, \mathbf{X} \rangle - \langle \mathbf{w}, \mathbf{\mu}_{\mid x_i} \rangle \qquad  \text{and} \qquad \sigma_N = \sqrt{\mathbf{w}^\top \Sigma_{\mid x_i}^{1} \mathbf{w}}~,
\end{equation*}
where $\mathbf{w}, \mathbf{X}$ are $N$-dimensional vectors, $\mathbf{\mu}_{\mid x_i} = \E[\mathbf{X} \mid X_i = x_i]$ is the vector of conditional means, and $\Sigma_{\mid x_i}^{1} \triangleq \text{Cov}[\mathbf{X} \mid X_i = x_i, Y = 1] = \Sigma_1 - \sigma_i^1 \sigma_i^{1\top}$.
Since $\sigma_N$ and $\mathbf{\mu}_{\mid x_i}$ are constant, we can write
\begin{align*}
    \PR_{\mathbf{X} \mid X_i = x_i, Y = 1} \left[  \langle \mathbf{w}, \mathbf{X} \rangle \geq 0 \right]
    &= \PR_{\mathbf{X} \mid X_i, Y = 1} \left[ \frac{ \langle \mathbf{w}, \mathbf{X} \rangle - \langle \mathbf{w}, \mathbf{\mu}_{j \mid x_i} \rangle }{ \sigma_N } \geq -\frac{\langle \mathbf{w}, \mathbf{\mu}_{j \mid x_i} \rangle }{ \sigma_N } \right] \\
    &= \PR_{Z \sim \NN(0,1)} \left[ Z \geq -\frac{\langle \mathbf{w}, \mathbf{\mu}_{j \mid x_i} \rangle }{ \sigma_N } \right] + o_N(1) \\
    &= 1 - \frac{1}{2} \left[ 1 + \operatorname{erf}\left( -\frac{\langle \mathbf{w}, \mathbf{\mu}_{j \mid x_i} \rangle }{ \sqrt{2} \sigma_N } \right) \right] + o_N(1) \\
    &= \frac{1}{2} + \frac{1}{2} \operatorname{erf}\left(\frac{\langle \mathbf{w}, \mathbf{\mu}_{j \mid x_i} \rangle }{ \sqrt{2} \sigma_N } \right) + o_N(1)~,
\end{align*}
where the second step, in which we acquire $o_N(1)$, follows from the definition of convergence in distribution.
Under Assumptions \labelcref{item:mean-assumption,item:covariance-assumption}, we may express this as
\begin{align*}
    \PR_{\mathbf{X} \mid X_i = x_i, Y = 1} \left[  \langle \mathbf{w}, \mathbf{X} \rangle \geq 0 \right]
    &= \frac{1}{2} + \frac{1}{2} \operatorname{erf}\left( \frac{\langle \mathbf{w}, x_i \sigma_i^y \rangle }{ \sqrt{2} \sqrt{ \mathbf{w}^\top \Sigma_1 \mathbf{w} - (\langle \mathbf{w}, \sigma_i^y \rangle)^2 } } \right) + o_N(1)~.
\end{align*}
Therefore,
\begin{align*}
    &\E_{\mathbf{X} \mid Y=1}[ \mathbbm{1}( \langle \mathbf{w}, \mathbf{X} \rangle \geq 0 ) X_i ] \\
    &= \E_{ X_i \mid Y=1 } \left[ X_i \left( \frac{1}{2} + \frac{1}{2} \operatorname{erf}\left( \frac{X_i}{\sqrt{2}} \cdot \frac{\langle \mathbf{w}, \sigma_i^y \rangle }{ \sqrt{ \mathbf{w}^\top \Sigma_1 \mathbf{w} - (\langle \mathbf{w}, \sigma_i^y \rangle)^2 } } \right) + o_N(1) \right) \right] \\
    &= \frac{1}{2} \E_{ X_i \mid Y=1 } \left[ X_i \operatorname{erf}\left( \frac{X_i}{\sqrt{2}} \cdot \frac{\langle \mathbf{w}, \sigma_i^y \rangle / \sqrt{\mathbf{w}^\top \Sigma_1 \mathbf{w}} }{ \sqrt{ 1 - (\langle \mathbf{w}, \sigma_i^y \rangle / \sqrt{\mathbf{w}^\top \Sigma_1 \mathbf{w}})^2 } } \right) \right] + \E_{X_i}[|X_i|] o_N(1) \\
    &= \frac{1}{2} \E_{ X_i \mid Y=1 } \left[ X_i \operatorname{erf}\left( \frac{X_i}{\sqrt{2}} \cdot \operatorname{alg}^{-1} \left( \frac{ \langle \mathbf{w}, \sigma_i^y \rangle }{ \sqrt{\mathbf{w}^\top \Sigma_1 \mathbf{w}} } \right) \right) \right] + \E_{X_i}[|X_i|] o_N(1)~.
\end{align*}
Defining $\varphi_i(a) \triangleq \E_{X_i \mid Y=1}[ X_i \operatorname{erf}(X_i \operatorname{alg}^{-1}(a) / \sqrt{2}) ]$ we can write
\begin{align*}
    \E_{\mathbf{X} \mid Y=1}[ \mathbbm{1}( \langle \mathbf{w}, \mathbf{X} \rangle \geq 0 ) X_i ]
    &= \frac{1}{2} \varphi_i \left( \frac{ \langle \mathbf{w}, \sigma_i^y \rangle }{ \sqrt{\mathbf{w}^\top \Sigma_1 \mathbf{w}} } \right) + \E_{X_i}[|X_i|] o_N(1)~.
\end{align*}
Note that $\E_{X_i}[|X_i|] \leq \sqrt{ \E_{X_i}[X_i^2] } = \sigma$ by Cauchy-Schwarz.
So, $\E_{X_i}[|X_i|] o_N(1) = o_N(1)$.
Moreover, by translation-invariance, all $X_i$ have the same marginal, so $\varphi_i \equiv \varphi_1 \triangleq \varphi$.
Thus,
\begin{align*}
    \E_{\mathbf{X} \mid Y=1}[ \mathbbm{1}( \langle \mathbf{w}, \mathbf{X} \rangle \geq 0 ) X_i ]
    &= \frac{1}{2} \varphi \left( \frac{ \langle \mathbf{w}, \sigma_i^y \rangle }{ \sqrt{\mathbf{w}^\top \Sigma_1 \mathbf{w}} } \right) + o_N(1)~.
\end{align*}
This form holds for all entries $i$.
Concatenating them, we obtain
\begin{align*}
    \E_{\mathbf{X} \mid Y=1}[ \mathbbm{1}( \langle \mathbf{w}, \mathbf{X} \rangle \geq 0 ) \mathbf{X} ]
    &= \frac{1}{2} \varphi \left( \frac{ \Sigma_1 \mathbf{w} }{ \sqrt{\mathbf{w}^\top \Sigma_1 \mathbf{w}} } \right) + o_N(1)~,
\end{align*}
which gives the desired result.
\qed

\subsection{Proof of lemma \ref{lem:varphi}} \label{subsec:pf_of_varphi}
Property 1 follows from the fact that $\operatorname{alg}$ and $\operatorname{erf}$ are odd functions.
This implies Property 4 since an odd function must have zero for even Taylor coefficients.

Properties 2 and 3 follow from differentiating \cref{eq:varphi}.
The first derivative is given by
\begin{align*}
    \varphi'(a) &= \frac{\sqrt{2}}{{\sqrt{{\pi}}}(1-a^2)^\frac{3}{2}} \E_{X_1}\left[ X_1^2\mathrm{e}^{-\frac{X_1^2a^2}{2(1-a^2)}} \right].
\end{align*}
Setting $a = 0$, we get
\begin{align*}
    \varphi'(0) 
    &= \sqrt{\frac{2}{\pi}} \E\left[ X_1^2 \right]
    = \sqrt{\frac{2}{\pi}} \sigma^2.
\end{align*}
The third derivative is given by
\begin{align*}
    \varphi'''(a) &= \frac{\sqrt{2}}{\sqrt{{\pi}}(1-a^2)^\frac{7}{2}(a^2-1)^2} \E_{X_1} \left[
    X_1^2(12a^6+(9X_1^2-21)a^4+(X_1^4-8X_1^2+6)a^2-X_1^2+3)\mathrm{e}^{-\frac{X_1^2a^2}{2(1-a^2)}}
    \right].
\end{align*}
Again setting $a = 0$ gives
\begin{align*}
    \varphi'''(0) &= \sqrt{\frac{2}{\pi}} \E_{X_1} \left[ X_1^2(-X_1^2+3) \right]
    = \sqrt{\frac{2}{\pi}} \left( 3 \E_{X_1}[X_1^2] - \E_{X_1}[X_1^4] \right) 
    = -\sqrt{\frac{2}{\pi}} (\kappa^4 \sigma^4 - 3 \sigma^2)~.
\end{align*}
\qed

\subsection{Proof of Proposition~\ref{thm:elliptical}}
The pdf of $X \sim \EE_N(\mu, \Sigma, \phi)$ is
\begin{align*}
    p_X(x) &= \frac{1}{\sqrt{\det(\Sigma)}} g( (x-\mu)^\top \Sigma^{-1} (x-\mu) )~,
\end{align*}
for some function $g : \R_{\geq 0} \to \R_{\geq 0}$ \cite{frahm2004generalized}.
A key property of elliptical distributions is that if $X \sim \EE_N(\mu, \Sigma, \phi)$, then its affine transformation is also elliptical: $\langle \mathbf{w}, \mathbf{X} \rangle \sim \EE_1(\langle \mathbf{w}, \mu \rangle, \mathbf{w}^\top \Sigma \mathbf{w}, \phi)$ for any $\mathbf{w} \in \R^N$.
Thus,
\begin{align*}
  p_{\langle \mathbf{w}, \mathbf{X} \rangle}(s) &= \frac{1}{\sqrt{\mathbf{w}^\top \Sigma \mathbf{w}}} \tilde{g}\left( \frac{(s-\langle \mathbf{w}, \mu \rangle)^2}{\mathbf{w}^\top \Sigma \mathbf{w}} \right),
\end{align*}
for some other function $\tilde{g} : \R_{\geq 0} \to \R_{\geq 0}$.

From our assumption of sign-symmetry, we have $\mu = 0$.
For brevity, we define $\sigma^2 \triangleq \mathbf{w}^\top \Sigma \mathbf{w}$ and $S \triangleq \langle \mathbf{w}, \mathbf{X} \rangle$.
We begin by computing (I) in \cref{eq:loss_2relu_neuron}.
Recall that we have $y = 0,1$ (\ie $K = 2$).
So,
\begin{align*}
    2 \times (I) 
    &= \E_{S \mid Y=1}[ \operatorname{ReLU}( S ) ]
    = \int_0^{\infty} \frac{1}{\sigma} \tilde{g}\left( \frac{s^2}{\sigma^2} \right) s \ \mathrm{d}s.
\end{align*}
At this point, we apply a $u$-substitution with $u = s^2 / \sigma^2$, and thus $\mathrm{d}u = 2 s \ \mathrm{d}s / \sigma^2 \iff \sigma \mathrm{d}u / 2 = s\ \mathrm{d}s / \sigma$.
This yields
\begin{align*}
    2 \times (I) 
    &= \frac{\sigma}{2} \underbrace{\int_0^{\infty} \tilde{g}(u)\ \mathrm{d}u}_{\triangleq C}
    = \frac{C}{2} \sqrt{\mathbf{w}^\top \Sigma_1 \mathbf{w}}.
\end{align*}
Recall that we assume the MSE loss $\LL$ is finite.
The first term in \cref{eq:loss_2relu_neuron} is clearly finite for $y = 0,1$.
The term $(II)$ is also easily seen to be finite, since it evaluates to $\mathbf{w}^\top (\Sigma_0 + \Sigma_1) \mathbf{w} / 4$.
Thus, $\LL$ being finite implies $(I)$ in \cref{eq:loss_2relu_neuron} is finite,  which implies $C < \infty$.
We computed (II) from \cref{eq:loss_2relu_neuron} above as
\begin{align*}
  \frac{1}{4} \mathbf{w}^\top \left( \Sigma_0 + \Sigma_1 \right) \mathbf{w}
\end{align*}
for $K = 2$.
Plugging in (I) and (II) and differentiating with respect to $\mathbf{w}$, we get
\begin{equation}
  \frac{1}{\tau} \frac{\mathrm{d}\mathbf{w}}{\mathrm{d}t} = \frac{\Sigma_1 \mathbf{w}}{\sqrt{\mathbf{w}^\top \Sigma_1 \mathbf{w}}} - \frac{1}{2} \left( \Sigma_0 + \Sigma_1 \right) \mathbf{w}~. \label{eq:elliptical_gradient_flow}
\end{equation}
The steady states of \cref{eq:elliptical_gradient_flow} thus satisfy
\begin{align*}
  C \frac{\Sigma_1 \mathbf{w}}{\sqrt{\mathbf{w}^\top \Sigma_1 \mathbf{w}}} 
  &= \frac{1}{2} \left( \Sigma_0 + \Sigma_1 \right) \mathbf{w}~.
\end{align*}
Recall that translation-invariance implies that $\Sigma_0$ and $\Sigma_1$ are circulant, and since they are covariance matrices, they are symmetric.
Then, they diagonalize in the basis given by the real and imaginary parts of the first $n/2$ Fourier components in the discrete Fourier transform, which we denote by the $n \times n$ real matrix $\FF$.
Note that $\FF$ is orthogonal.
Define $\mathbf{v} = \FF^\top \mathbf{w}$ and $\Lambda_y = \FF^\top \Sigma_y \FF$ for $y = 0,1$.
Thus, the steady states satisfy
\begin{align*}
  C \frac{\Lambda_1 \mathbf{v}}{\sqrt{\mathbf{v}^\top \Lambda_1 \mathbf{v}}} &= \frac{1}{2} \left( \Lambda_0 + \Lambda_1 \right) \mathbf{v}~.
\end{align*}
This holds iff for all $i \in [n]$,
\begin{align*}
  C (\Lambda_1)_{ii} (\mathbf{v}^\top \Lambda_1 \mathbf{v})^{-\frac{1}{2}} v_i &= \frac{1}{2} ( (\Lambda_0)_{ii} + (\Lambda_1)_{ii} ) v_i~.
\end{align*}
Thus, if $v_i \neq 0$, we must have that
\begin{align*}
  \frac{(\Lambda_0)_{ii}}{(\Lambda_1)_{ii}} &= 2 C (\mathbf{v}^\top \Lambda_1 \mathbf{v})^{-\frac{1}{2}} - 1~.
\end{align*}
That is, the ratio of the $i$-th eigenvalues of $\Sigma_0$ and $\Sigma_1$ must be constant for all $i$ s.t. $v_i \neq 0$.
The eigenvalues of these matrices always come in pairs because of how we defined $\FF$ using both the real and imaginary parts of the discrete Fourier transform.
In general, we observe that each pair assumes a unique value.
So, since $C$ is finite, the condition above can hold for at most two distinct values of $i$.
Therefore, $v_i = 0$ for all but at most two $i \in [n]$, implying that the steady state $w$ is of the form $a \cos(2\pi k x) + b \sin(2 \pi k x)$, \ie it is oscillatory.
As such, it is \emph{not} localized. 
\qed

\section{Additional experiments}

\subsection{Visualizing breakdown of Assumption~\labelcref{item:lindeberg-condition}}

\begin{figure}[htbp]
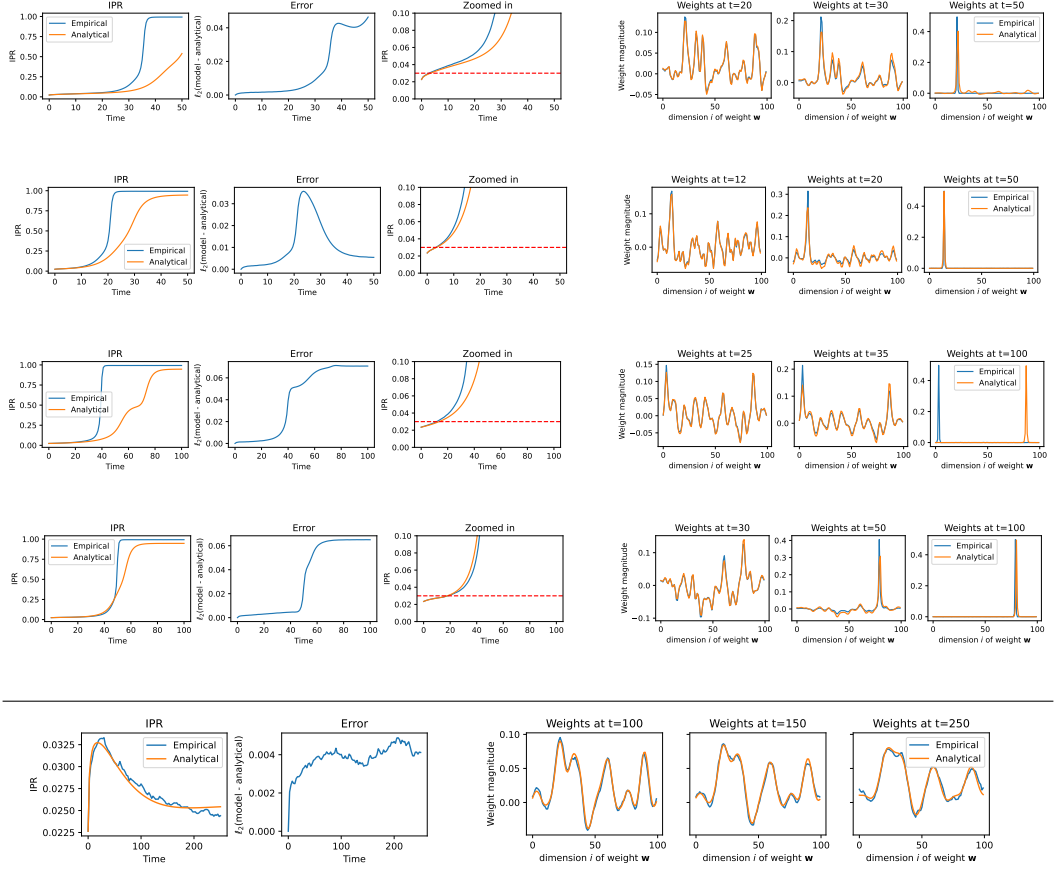

        \centering
            \vspace{-16pt}
        \foreach \row in {0,1,2,3}{
            \foreach \col in {ipr_mse_zoomed_in, timeshot}{
                \includegraphics[height=50pt]{rebuttal-figures/time/seed=\row/\col.pdf}
                \hspace{10pt}
            }
            \\ 
            \foreach \col in {ipr_mse_zoomed_in, timeshot}{
                \label{fig:\row\col}
            }
            \vspace{4pt} 
        }
        \vspace{4pt}
        \hrule
        \vspace{4pt}
        \foreach \col in {ipr_mse, timeshot}{
            \includegraphics[height=60pt]{rebuttal-figures/time/seed=0_gaussian/\col.pdf}
            \hspace{10pt}
        }
        \\ 
        \foreach \col in {ipr_mse, timeshot}{
            \label{fig:gaussian_\col}
        }
        \caption{
(\textbf{Top}) Four initializations trained on $\texttt{NLGP}(g=100)$ with $\xi_0 = 0.3$ and $\xi_1 = 0.7$.
As expected, weights always localize.
In (Left, First) we plot IPR for empirical and analytical receptive fields (RFs) across time (defined as (\# of gradient steps) $\times \ \, \tau$, the learning rate).
In (Left, Second) we plot the time-evolution of $\ell_2$ distance between the empirical and analytical RFs.
In (Left, Third) we zoom in on (Left, First), restricting the range to $[0,0.1]$ to more closely see divergence in IPR early in training.
In (Right, First) and (Right, Second), we snapshot the empirical and analytical RFs at a time \emph{before} and \emph{just after}, respectively, the analytical model breaks down  (according to IPR and $\ell_2$ distance) due to localization.
Finally, in (Right, Third), we snapshot \emph{at the end} of the training period.
(\textbf{Bottom}) Same initialization as first row in \textbf{top}, but trained on $\texttt{NLGP}(g=0.01)$ data, again with $\xi_0 = 0.3$ and $\xi_1 = 0.7$.
As expected, weights do not localize.
We plot the same quantities as above, but here the predictions of our analytical model hold \emph{throughout} the entire training process as localization never emerges and so assumption (A3) is not violated as above.
\label{fig:time}
}
    \end{figure}

\cref{fig:time} demonstrates that our analytical model holds for long enough during training to capture the emergence of localization in the single ReLU neuron (\labelcref{item:single-neuron-model}). 
In the first three columns, we visualize the IPR of the weights from our empirical and analytical models, as well as the $\ell_2$ difference between these two weights.
In the first four rows, we visualize these metrics for four random initializations of the model, training each on $\texttt{NLGP}(g=100)$ with $\xi_0 = 0.3$ and $\xi_1 = 0.7$, where we expect from \cref{thm:localization} to see localization. 
We see the error rapidly increase shortly after IPR increases, indicating the formation of localized receptive fields.
The last three columns confirm this, as they show snapshots from \emph{before}, \emph{during}, and \emph{after} the divergence between the empirical and analytical weights.
We observe the weights are nearly identical \emph{before}, differ only slightly at the most localized point \emph{during}, and are both localized \emph{after}, but possibly with different magnitudes and positions.
The difference that emerges \emph{during} is due to a breakdown of Assumption~\labelcref{item:lindeberg-condition} used to create our analytical model, which is violated when the norm of $\mathbf{w}$ is dominated by just a few entries, \ie it is localized.
While \cite{ingrosso2022data} also observe a breakdown in their analytical model as localization emerges, ours, crucially, holds for long enough to characterize the emergence of localization.

We discuss the individual subplots in more detail.
In all but the third row of \cref{fig:time}, the analytical predictions are near-exact; in the third row, we predict localization, but at the wrong position.
Focusing again on the first row, we see that at $t=20$, the weights have not yet become localized (from IPR, left, first, and visually) and analytical and empirical weights match nearlt exactly, as confirmed by the small distance in left, center above. 
At $t=30$, a localized peak around $i=21$ begins to emerge, violating Assumption \labelcref{item:lindeberg-condition} and weakening analytical precision.
The analytical model then underestimates the degree to which the main peak at $i=21$ dominates, while it overestimates the size of competing peaks at $i=30$, $37$, and $90$. 
Despite this, at $t=50$, we see that predictions from the analytical model retain a match to the empirical model. 

In the last row of \cref{fig:time}, we use the same initialization and setting as in the first row, except that we train on $\texttt{NLGP}(g=0.01)$ data instead.
From \cref{thm:localization}, we \emph{do not} expect to see localization.
The evolution of IPR confirms this, as it stays low in magnitude.
We also see that, because localization never emerges, Assumption~\labelcref{item:lindeberg-condition} is never violated, and so our analytical model accounts for the empirical model nearly perfectly.

\end{document}